\documentclass[11pt]{article}

\usepackage[preprint]{acl}
\usepackage{times}
\usepackage{latexsym}
\usepackage[T1]{fontenc}
\usepackage[utf8]{inputenc}
\usepackage{microtype}
\usepackage{graphicx}

\usepackage{amsmath, amssymb}
\usepackage{booktabs}
\usepackage{makecell}
\usepackage{tcolorbox}

\newcounter{takeaway}
\newtcolorbox{takeaway}{colback=blue!5!white, colframe=blue!55!black,
  boxrule=0.6pt, arc=4pt, left=5pt, right=5pt, top=3pt, bottom=3pt,
  before skip=6pt, after skip=6pt,
  before upper={\refstepcounter{takeaway}\textit{Takeaway~\thetakeaway:}\ }}

\newcommand{\ci}[1]{{\scriptsize$#1$}}

\ifdefined\pdfminorversion \pdfminorversion=7 \fi

\title{Same Answer, Different Confidence:\\
        Protocol Sensitivity in Measuring LLM Confidence}

\author{
  Hankyeol Kim \and Pilsung Kang\textsuperscript{*} \\
  Seoul National University \\
  \texttt{\{hankyeol,pilsung\_kang\}@snu.ac.kr} \\
  \textsuperscript{*}Corresponding author
}

\begin{document}
\maketitle


\begin{abstract}
Large language models (LLM) can state their confidence in an answer
(verbalized confidence), or the confidence can be read from the probability of
the answer's tokens (token confidence).
Comparisons of the two ask which is better calibrated, but the token
probability of an answer depends on the text that precedes it.  Across five open LLMs and four
question-answering datasets, we fix each answer, its verbalized confidence,
and its correctness label, and vary how each score is measured.  The scores
themselves change with the protocol, and so does the comparison between them.
Changing
the context in which the token probability is read shifts the gap between the
two scores by about the size of the gap itself ($.287$ vs.\ $.304$ in expected
calibration error, ECE; $.189$ vs.\ $.145$ in AUROC).  The 95\% interval of the
shift excludes zero in 19 and 17 of the 20 settings, and under AUROC the shift
has the same sign in all of them, so the context changes how answers are
ranked.  The better score flips in 8 settings under each metric, counting only
flips that the intervals support, and the shift also appears in four larger
models of up to 72B parameters.  Most of the shift appears with changes to
the instruction, not to the position of the answer.  Other measurement choices
also matter, and stated confidence depends on how a supplied answer is
written.  We turn these results into a reporting
checklist for LLM confidence and its comparisons.
\end{abstract}


\section{Introduction}
\label{sec:intro}

\begin{figure*}[t]
\centering
\includegraphics[width=\textwidth]{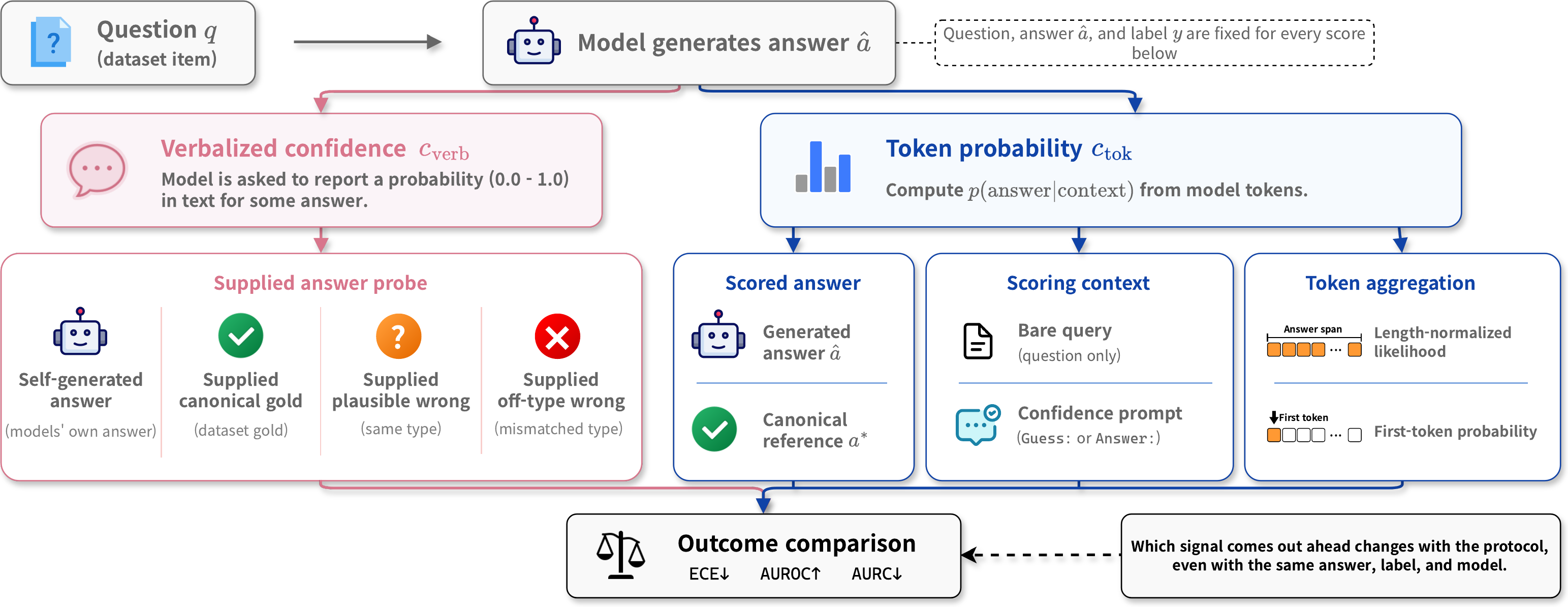}
\caption{Overview of the comparison.  The question, the model's answer $\hat
a$, and its label $y$ stay fixed.  The token score varies with the scoring
context and the token aggregation (the length-normalized likelihood is the
geometric mean of the answer-token probabilities).  Scoring the canonical
reference breaks the alignment rule (\S\ref{sec:events}) and is used only as a
diagnostic.  The supplied-answer probe measures verbalized confidence for
answers we give the model.  Table~\ref{tab:axes} lists all choices we vary.}
\label{fig:overview}
\end{figure*}

A language model's confidence in an answer can be measured in two ways.  The
model can state a probability in words, or we can compute the probability of
the answer's tokens under the model.  Studies that compare these verbalized and
token scores usually ask which of the two methods is better calibrated, and
they reach different conclusions \citep{Tian2023,Ni2024}.

The token score of an answer depends on the text that precedes it: the same
answer gets a different token score after a bare question than after the
prompt that asked the model for its answer and its confidence.  Prior work has
studied how each score changes with the prompt or the metric
(\S\ref{sec:rel-sensitivity}), but less is known about how such choices change
which of the two scores looks better.

Token probabilities are known to vary with the prompt \citep{Zhao2021}.
However, it remains unclear how much each score changes with the measurement
protocol, and whether the change is large enough to alter the conclusion of a
comparison between them.  To measure
this, we fix each model's answer, its verbalized score, and its correctness
label, and vary the remaining measurement choices
(Figure~\ref{fig:overview}).  Most importantly, we read the token score of the
same answer in three contexts.  We also vary how token probabilities are
combined, the correctness rule, and the estimator of expected calibration
error (ECE; \S\ref{sec:metrics}), and we probe the
verbalized score with answers that we supply.  We test five open
instruction-tuned models on four question-answering datasets, which gives 20
settings, and check four larger models of up to 72B parameters.

Changing the context from a bare question to the prompt that produced the
answer changes the gap between the two scores by about as much as the gap
itself.
Under ECE, the gap moves by $.287$ on average, against a bare-context gap of
$.304$; under AUROC, which measures how well a score ranks right answers
above wrong ones, it moves by $.189$, against $.145$.  Under AUROC, the
shift goes the same way in all 20 settings, so the context changes the order of
the answers, not only the scale of the scores.  We show this shift also appears in the larger models.  Most of the shift appears when we change the instruction in the prompt, not
where the answer begins.  As a result, two prompts that both omit a confidence
request can still lead to different conclusions.  Other choices, such as the
correctness rule and the way token probabilities are combined, also change the
gap, in some settings enough to change which score looks better.  On the verbalized side, stated confidence
for a supplied answer changes with how a correct answer is written, barely
separates a correct answer from a plausible wrong one, and leans toward the
model's own wrong answers.

Our contributions are as follows:
\begin{itemize}
\setlength\itemsep{0pt}
\item We study the sensitivity and robustness of LLM confidence across
  measurement choices, a prompt ablation, and supplied answers
  (\S\ref{sec:results}).
\item Using a design that fixes the answer and its label, we show that the
  context in which the token score is read changes the comparison between
  verbalized and token confidence about as much as the gap itself
  (\S\ref{sec:method}, \S\ref{sec:res-context}).
\item We identify the factors that matter most for measuring LLM confidence
  and summarize them as a reporting checklist (Table~\ref{tab:checklist}).
\end{itemize}


\section{Related Work}
\label{sec:related}

\subsection{Verbalized and Token Confidence}
\label{sec:rel-signals}

A language model can state its confidence in words or numbers
\citep{LinHiltonEvans2022,Mielke2022}.  It can give a probability together with
its answer \citep{Tian2023}, or rate a proposed answer, as in P(True)
\citep{Kadavath2022}.  Token confidence instead reads the probability that the
model assigns to an answer, as a single-token probability, a sequence
likelihood, a length-normalized likelihood, or an inverse perplexity
\citep{DesaiDurrett2020,Lyu2024,Huang2024,Khanmohammadi2025}.  Studies that
compare the two reach different conclusions.  \citet{Tian2023} find that
verbalized scores are better calibrated than sampled conditional
probabilities, whereas \citet{Ni2024} find token confidence better for less
common questions and \citet{Xiong2024} find agreement among sampled answers
better than verbalized confidence.  We show that which signal looks better can
change with how the comparison is measured, even for the same answers.

\subsection{Sensitivity of Confidence Scores to Measurement}
\label{sec:rel-sensitivity}

Both signals depend on how they are measured.  Verbalized confidence changes
with the prompt \citep{Xiong2024,Yang2024} and the response scale
\citep{Dai2026}, and token confidence with how token probabilities are combined
\citep{Yang2024}.  ECE (\S\ref{sec:metrics}) changes with its estimator
\citep{Nixon2019,BlasiokNakkiran2024}, and ECE alone can mislead when methods
are compared \citep{Muller2026,Wu2026}.  Studies that change the prompt usually
also change the answer being scored.

\begin{table*}[t]
\footnotesize
\centering
\setlength{\tabcolsep}{6pt}
\begin{tabular}{@{}llp{0.5\textwidth}@{}}
\toprule
Axis & Side & Values compared \\
\midrule
Context and answer slot & token &
  bare context; \texttt{Answer:} slot; \texttt{Guess:} slot \\
Token aggregation & token &
  length-normalized likelihood; first-token probability \\
Correctness rule and answer format & shared &
  normalized containment; exact match; multiple-choice format \\
Calibration estimator & shared &
  10-bin ECE; KDE; KS; Smoothed ECE \\
\midrule
\multicolumn{3}{@{}l}{\emph{Probe of verbalized confidence, not a protocol
  axis} (\S\ref{sec:supplied}):} \\
Supplied answer & verbalized &
  own; canonical reference; plausible wrong; off-type \\
\midrule
\multicolumn{3}{@{}l}{\emph{Fixed by the alignment rule, not an
  axis:} scored answer $=$ rated answer (\S\ref{sec:events}).} \\
\bottomrule
\end{tabular}
\caption{The axes of a verbalized-versus-token comparison, and the
supplied-answer probe.  The probe measures the level of verbalized confidence,
not the gap.}
\label{tab:axes}
\end{table*}

\subsection{Scoring Context and Rated Answer}
\label{sec:rel-position}

In this study, we change where the token score is read while keeping the
question, the answer, and its label the same.  The closest design is that of
\citet{Xia2026}, who score the same answer under several formulations of the
task prompt, but not in contexts that differ in whether they ask for
confidence.  Prior work differs in where it reads the token score.
\citet{Huang2024} and \citet{Kumar2024} read it under the prompt that produced
the answer, whereas \citet{Ni2024} and \citet{Tao2025} read it under a prompt
that also asks for confidence.  Others, such as \citet{Lyu2024}, do not say
where (Appendix~\ref{app:audit}).

Verbalized confidence also depends on which answer the model rates.  A chat
model is more confident in its own answer than in the same answer attributed to
the user \citep{SanzGuerrero2026}, and its stated confidence can be largely
independent of the answer \citep{Seo2026}.  Our supplied-answer probe shows
related patterns without attributing answers to anyone
(\S\ref{sec:res-verbal}).


\section{Method}
\label{sec:method}

Comparing verbalized and token confidence requires several choices, and each
can change the result.  We first define what both scores must describe and
list the choices, then describe our setup.

\subsection{Confidence Scores and Alignment}
\label{sec:events}

A confidence score expresses how likely a model's answer $\hat a_i$ to a
question $q_i$ is correct.  We compare two such scores for the same answer.
The \emph{verbalized score} $c_{\mathrm{verb},i}$ is the probability that the
model states for $\hat a_i$.  The \emph{token score} $c_{\mathrm{tok},i}$ is
instead read from the model's output distribution.  It is the geometric mean of
the probabilities that the model assigns to the tokens of $\hat a_i$ after some
input text (forced decoding), so it does not depend on the answer's length.  Both scores are evaluated against the correctness label $y_i$,
which records whether $\hat a_i$ is correct.

\paragraph{Alignment rule.}  The verbalized score, the token score, and the
correctness label must all refer to the same model answer to the same
question.  Scoring the reference
answer instead of the model's own answer breaks this rule
(\S\ref{sec:res-ledger} reports what it changes).

\subsection{Protocol Choices}
\label{sec:axes}

We study how the measurement protocol changes the measured confidence of a
model, and the comparison between its two confidence signals.  A
\emph{protocol} is a full set of choices, and we call each choice an
\emph{axis}.  Table~\ref{tab:axes} lists the axes we vary.

The most important axis is the scoring context.  The \emph{scoring context},
or \emph{context} for short, is the text placed before the answer when the token
score is read, and the \emph{answer slot} is where the answer starts.  The answer was generated in only one context.  In
every other context, the token score describes an answer the model did not
produce there.

\subsection{Models and Data}
\label{sec:setup}

We use five open instruction-tuned models from three developers:
Llama-3-8B-Instruct \citep{Llama3}, Mistral-7B-Instruct-v0.3
\citep{Mistral7B}, Qwen2.5-7B-Instruct \citep{Qwen25}, Qwen3.5-9B
\citep{Qwen35} without thinking mode, and Ministral-3-8B-Instruct
\citep{Ministral3}, each with its released chat template.  We use TriviaQA
\citep{Joshi2017}, SciQ \citep{Welbl2017}, TruthfulQA \citep{Lin2022}, and
MMLU \citep{Hendrycks2021} with four subjects, which gives five models
$\times$ four datasets $=$ 20 \emph{settings}, each with 734 to 1,155
items.  Answers use greedy decoding, and we drop unparsed
rows; the lowest parse rate is $.898$.  We also repeat the main analysis on
larger models (Qwen2.5-14B, 32B, and 72B, and Ministral-3-14B) and on the base
versions of the five models.  Details are in Appendix~\ref{app:repro}.  We release code, prompts, labels, and outputs.

\subsection{Contexts and Correctness}
\label{sec:contexts}

\paragraph{Contexts.}  We fix each model's answer and its verbalized score, and
read the token score of the same answer in three contexts
(Figure~\ref{fig:contexts}).  Both the answer and the verbalized score come from
one generation under the confidence prompt of \citet{Tian2023}.  The prompt asks
for ``your best guess and the probability that it is correct (0.0 to 1.0)'',
with only these two as output: the guess, written after \texttt{Guess:}, is the
model's answer $\hat a_i$, and the probability, written after
\texttt{Probability:}, is how likely the model says that guess is correct,
which is the verbalized score.  Each model's chat template splits the input
into a \emph{user turn}, which holds the prompt, and an \emph{assistant turn},
where the model writes its reply.  The three contexts differ in the text before
$\hat a_i$:
\begin{itemize}
\setlength\itemsep{0pt}
\item \emph{Bare context}: the user turn is the question followed by
  \texttt{Answer:}; $\hat a_i$ starts the assistant turn.
\item \emph{\texttt{Answer:} slot}: the user turn is the question alone; the
  assistant turn starts with \texttt{Answer:}, followed by $\hat a_i$.
\item \emph{\texttt{Guess:} slot}: the user turn is the confidence prompt; the
  assistant turn starts with \texttt{Guess:}, followed by $\hat a_i$.  This is
  where the model generated $\hat a_i$.
\end{itemize}
The first two contexts differ only in where the turn-switch tokens of the chat
template fall relative to \texttt{Answer:} and, as a result, in whether the
answer starts with a space.  The third also changes the
instruction, the marker, and the generation context.

\begin{figure}[t]
\scriptsize
\centering
\newcommand{\given}[1]{\colorbox{black!14}{\strut #1}}
\newcommand{\scored}[1]{\colorbox{blue!22}{\strut\texttt{#1}}}
\setlength{\tabcolsep}{2pt}
\renewcommand{\arraystretch}{1.2}
\begin{tabular}{@{}p{0.2\columnwidth}@{\hspace{1.6em}}p{0.3\columnwidth}p{0.3\columnwidth}@{}}
\toprule
Context & User turn & Assistant turn \\
\midrule
Bare &
  \given{\texttt{\{q\}}} \given{\texttt{Answer:}} &
  \scored{\{a\}} \\
\texttt{Answer:} slot &
  \given{\texttt{\{q\}}} &
  \given{\texttt{Answer:}} \scored{\{a\}} \\
\texttt{Guess:} slot &
  \given{confidence prompt} &
  \given{\texttt{Guess:}} \scored{\{a\}} \\
\midrule
\multicolumn{3}{@{}l}{\emph{Prompt ablation only}} \\
Answer-only &
  \given{answer-only instr.} &
  \given{\texttt{Answer:}} \scored{\{a\}} \\
\bottomrule
\end{tabular}
\caption{Scoring contexts for one fixed answer (\texttt{\{a\}}).
\colorbox{black!14}{Gray}: text given to the model as input; the prompts
contain the question \texttt{\{q\}}.  \colorbox{blue!22}{Blue}: the answer
tokens, whose probability is the token score.  The first three rows are the
three contexts; the last row is used only in the prompt ablation.}
\label{fig:contexts}
\end{figure}

\paragraph{Prompt ablation.}  The \texttt{Answer:} slot and the \texttt{Guess:}
slot differ in both the user prompt and the marker.  To see which of the two
moves the gap, we cross four user prompts with the markers \texttt{Answer:} and
\texttt{Guess:}.  The prompts add the parts of the confidence prompt one at a
time: the question alone; an \emph{answer-only instruction}, which asks only for
a short guess; the confidence prompt without its probability request; and the
full confidence prompt.  We call the answer-only instruction with
\texttt{Answer:} the \emph{answer-only context} (Appendix~\ref{app:ablation}).

\paragraph{Answers from another prompt.}  In the bare context and the
\texttt{Answer:} slot, the token score describes an answer that the model
generated under another prompt.  To test whether this matters, we compare
answers to the same question that come from two prompts.  Besides the answers
$\hat a_i$ from the confidence prompt, we also ask each model the question under
the answer-only instruction, which gives a second set of answers $\hat b_i$.  We
score both sets in the same contexts, on the items that have both.

\paragraph{Correctness.}  The correctness label $y_i$ depends on the
\emph{correctness rule} that grades the answer.  Our default is normalized
containment: after normalization, the answer equals a gold answer or one of the
two strings contains the other (Appendix~\ref{app:repro}).  We also use exact
match and a multiple-choice format, in which the model sees the options of SciQ
and MMLU and answers with a letter (Appendix~\ref{app:correctness}).

\subsection{Metrics}
\label{sec:metrics}

We use three metrics, defined in Appendix~\ref{app:repro}.  ECE (expected
calibration error) is the average difference between confidence and accuracy
over confidence bins, 10 equal-width bins by default.  AUROC is the probability
that a correct answer receives higher confidence than a wrong one.  AURC is the
average error when answers are kept in order of confidence.  ECE reads a score as
a probability of correctness, which the token score need not be.  AUROC and AURC
use only the ranking of scores, so they do not depend on their scale.

For context $X$, the \emph{gap}
\[
g^M_X = M(c_{\mathrm{verb}},y) - M(c^X_{\mathrm{tok}},y)
\]
compares the two signals on the same answers, where $M$ is a metric and $X$
is the context in which the token score is read.  We write $g_{\mathrm{bare}}$,
$g_{\texttt{Answer:}}$, and $g_{\texttt{Guess:}}$ for the three contexts.  The
sign of $g$ depends on the metric.  For ECE and AURC, lower is better, so
$g>0$ means the token score is better.  For AUROC, higher is better, so $g>0$
means the verbalized score is better.  The \emph{context shift} is
$\Delta g = g_{\texttt{Guess:}} - g_{\mathrm{bare}}$.

We measure the size of the shift by the \emph{mean $|\Delta g|$} over the 20
settings, and its consistency by the number of settings whose 95\% paired
bootstrap interval of $\Delta g$ excludes zero.  We also count \emph{flips},
settings where the better signal differs between the bare context and the
\texttt{Guess:} slot.  A flip judged by point estimates may reflect
noise, so we call a flip \emph{fully resolved} when the 95\% intervals of $g$ in
both contexts exclude zero, and we give the point-estimate count in parentheses.
Because flips also depend on how close the two signals are, we treat them as
secondary to the shift itself.

\subsection{Supplied Answers}
\label{sec:supplied}

The main comparison always rates the model's own answer, so the source of the
rated answer is not a protocol axis.  We still vary it, to check how sensitive
and reliable the verbalized score is.  We ask whether it depends on how a
correct answer is written, whether it separates right from wrong answers, and
whether it favors the model's own answers.

We place a \emph{supplied answer} after \texttt{Guess:} and let the model write
only the probability.  The supplied answer is the canonical reference (the
dataset's reference answer) or a \emph{plausible wrong} answer (another model's
wrong answer to the same question).  When no other model answered the question
wrongly, we supply an \emph{off-type} answer instead (the reference answer of
another question), only as a check that confidence can drop.  The model is not
told who wrote a supplied answer.  Re-supplying the model's own answer changes
its confidence by at most $.022$ in any setting.  On TriviaQA, which lists
several accepted spellings of an answer, we also supply two correct forms of the
same answer, the first listed alias and the canonical reference, and compare the
confidence they receive (Appendix~\ref{app:provenance}).


\section{Results}
\label{sec:results}

\subsection{Context Shift}
\label{sec:res-context}

\begin{figure*}[t]
\centering
\includegraphics[width=\textwidth]{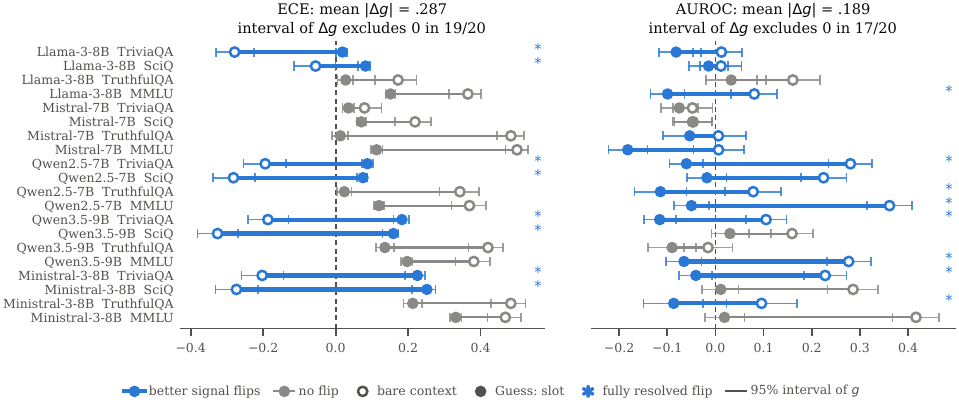}
\caption{The gap for the same answers in the bare context (open circle) and the
\texttt{Guess:} slot (filled circle), one row per setting, with 95\%
intervals.  Blue segments cross zero, so the better signal flips; an asterisk
marks a fully resolved flip.}
\label{fig:shift}
\end{figure*}

Changing the scoring context from the bare context to the \texttt{Guess:} slot
moves the gap about as much as the gap itself (Table~\ref{tab:main}).
Averaged over the 20 settings (\S\ref{sec:setup}), the gap
shifts by $.287$ under ECE, and the bare-context gap is $.304$.  Under AUROC,
the two numbers are $.189$ and $.145$.  For comparison, switching the ECE
estimator moves the gap by only $.005$ to $.027$ (\S\ref{sec:res-axes}).  The
shift is rarely noise, since its interval excludes zero in 19 of 20 settings
under ECE and in 17 under AUROC (Figure~\ref{fig:shift}).

\begin{table}[t]
\footnotesize
\centering
\setlength{\tabcolsep}{2.5pt}
\renewcommand{\arraystretch}{0.95}
\resizebox{\columnwidth}{!}{%
\begin{tabular}{@{}lccc@{}}
\toprule
20 settings & ECE $\downarrow$ & AUROC $\uparrow$ & AURC $\downarrow$ \\
\midrule
\multicolumn{4}{@{}l}{\emph{Shift, bare context $\to$ \texttt{Guess:} slot}} \\
\quad mean $|\Delta g|$ & $.287$ & $.189$ & $.106$ \\
\quad mean $|g_{\mathrm{bare}}|$, for scale & $.304$ & $.145$ & $.091$ \\
\quad interval of $\Delta g$ excludes 0 & 19/20 & 17/20 & 14/20 \\
\quad flips & 8 (8) & 8 (13) & 6 (7) \\
\midrule
\multicolumn{4}{@{}l}{\emph{Levels, mean over settings}} \\
\quad verbalized $c_{\mathrm{verb}}$ & $.433$ & $.702$ & $.444$ \\
\quad token, bare context               & $.309$ & $.568$ & $.496$ \\
\quad token, \texttt{Answer:} slot   & $.297$ & $.549$ & $.511$ \\
\quad token, \texttt{Guess:} slot    & $.308$ & $.757$ & $.392$ \\
\midrule
\multicolumn{4}{@{}l}{\emph{Gaps, mean over settings}} \\
\quad $g_{\mathrm{bare}}$   & $+.124$ & $+.134$ & $-.052$ \\
  & \ci{[+.111,+.134]} & \ci{[+.123,+.145]} & \ci{[-.060,-.045]} \\
\quad $g_{\texttt{Answer:}}$ & $+.137$ & $+.153$ & $-.067$ \\
  & \ci{[+.124,+.146]} & \ci{[+.142,+.164]} & \ci{[-.075,-.060]} \\
\quad $g_{\texttt{Guess:}}$ & $+.126$ & $-.055$ & $+.052$ \\
  & \ci{[+.120,+.129]} & \ci{[-.065,-.046]} & \ci{[+.045,+.058]} \\
\bottomrule
\end{tabular}
}
\caption{The same answers scored in three contexts.  Flips: fully resolved
(point estimate).  Brackets: 95\% paired bootstrap intervals.  Positive $g$
favors the token score under ECE and AURC and the verbalized score under
AUROC.}
\label{tab:main}
\end{table}

Under AUROC, the shift always goes the same way: in all 20 settings, the token
score ranks the answers better in the \texttt{Guess:} slot.  Its AUROC rises
from $.568$ to $.757$ on average, while the verbalized score stays at $.702$.
AUROC depends only on the order of the scores, so rescaling alone cannot
change it.  The rise means that the context changes the order of the answers.

Under ECE, the shift has no fixed direction.  The context also changes the
level of the token score: its mean rises from $.149$ in the bare context to
$.732$ in the \texttt{Guess:} slot.  In the bare context, the token score is
underconfident: its mean is below accuracy in 18 settings.  In the
\texttt{Guess:} slot, it is overconfident: its mean is above accuracy in all
20.  So in which context the token score is better calibrated depends on the
accuracy.  When
accuracy is above $.5$, the token score is better calibrated in
the \texttt{Guess:} slot in 8 of 10 settings.  When accuracy is below $.5$, it
never is.

The shift often flips which signal looks better (\S\ref{sec:metrics}).
Under ECE, the better signal flips in 8 settings, and all 8 flips are fully
resolved.  Under AUROC, 8 flips are fully resolved, and 5 more appear only in
the point estimates.

The shift also appears when the model states its confidence on a 0--100 scale
instead of 0.0--1.0 (Appendix~\ref{app:promptscale}).  It appears in larger
models as well (Appendix~\ref{app:larger}), and under exact match, a stricter
correctness rule (\S\ref{sec:res-correctness}).  It is weaker in base models, where
its interval excludes zero in 11 settings under ECE and 12 under AUROC
(Appendix~\ref{app:base}).

\begin{takeaway}
Where we read the token score can move the comparison as much as the gap we
are trying to measure.
\end{takeaway}

\subsection{Prompt Ablation and Baseline Choice}
\label{sec:res-ledger}

The gap moves mainly when the instruction changes, and little when
\texttt{Answer:} moves (Table~\ref{tab:ledger}).  Moving \texttt{Answer:} into
the assistant turn changes the AUROC gap only from $+.134$ to $+.153$.
Replacing the plain question with the answer-only instruction turns it
negative, to $-.019$.  The full confidence prompt then moves it a little
further, to $-.055$.  The ECE gap moves by $.031$ on average when
\texttt{Answer:} moves, but by $.175$ when the instruction changes
(Appendix~\ref{app:ablation}).

\begin{table}[t]
\footnotesize
\centering
\setlength{\tabcolsep}{2pt}
\renewcommand{\arraystretch}{0.95}
\resizebox{\columnwidth}{!}{%
\begin{tabular}{@{}llc@{}}
\toprule
Context & Change & AUROC $g$ \\
\midrule
Bare & -- & $+.134$ \ci{[+.123,+.145]} \\
\texttt{Answer:} slot & \texttt{Answer:} to assistant turn
  & $+.153$ \ci{[+.142,+.164]} \\
Answer-only & answer-only instruction & $-.019$ \ci{[-.029,-.009]} \\
\texttt{Guess:} slot & confidence prompt, \texttt{Guess:}
  & $-.055$ \ci{[-.065,-.046]} \\
\bottomrule
\end{tabular}
}
\caption{The AUROC gap for the same answers in four contexts, mean over 20
settings, with 95\% paired bootstrap intervals.  Each row makes one change
(Change) to the context above it.  Negative $g$ favors the token score.}
\label{tab:ledger}
\end{table}

Which prompt serves as the baseline matters even among prompts that do not
ask for confidence.  The \texttt{Answer:} slot and the answer-only context
both use such prompts, yet the better signal differs between them in 4
settings under ECE and 3 under AUROC (fully resolved flips).  The shift to the \texttt{Guess:} slot also
depends on this choice.  From the bare context, it is $.287$ under ECE and
$.189$ under AUROC (\S\ref{sec:res-context}).  From the answer-only context,
it is only $.113$ and $.050$, with no fully resolved flip.

In the bare context, the token score hardly favors answers from either
prompt.  In the \texttt{Guess:} slot, it favors the answer generated there.  For
each question, we compare two answers from the same model: one generated under
the confidence prompt and one generated under the answer-only instruction.  In
the bare context, the two answers get almost the same token score (they differ
by $.008$ on average).  In the \texttt{Guess:} slot, the answer generated there
gets a token score $.187$ higher (Appendix~\ref{app:prompt-slot}).

The reference answer shows the same pattern.  Scoring it instead of the
model's answer breaks the alignment rule (\S\ref{sec:events}).  In the
\texttt{Guess:} slot, this lowers the token score by $.458$ on average and moves
the ECE gap by $.224$.  In the bare context, the two changes are only $.048$ and
$.054$ (Appendix~\ref{app:goldstring}).

\begin{takeaway}
The comparison depends mostly on the instruction text, not on where the
answer starts.  A report should give the full prompt, not only whether it
asks for confidence.
\end{takeaway}

\subsection{Correctness Rule and Answer Format}
\label{sec:res-correctness}

The correctness rule moves the gap mainly in the bare context and under ECE.
We leave out TruthfulQA, where exact-match accuracy is below $.05$, and use the
other 15 settings.  Switching from the default rule to exact match changes the
ECE gap by $.181$ in the bare context but by only $.005$ in the \texttt{Guess:}
slot (Table~\ref{tab:axis-size}).  Under AUROC, the two changes are $.033$ and
$.027$.

The bare context and the \texttt{Guess:} slot differ because the token score
lies on different sides of accuracy in the two (\S\ref{sec:res-context}).  ECE
measures how far a score is from accuracy.  Under exact match, accuracy is
$.106$ lower on average.  In the bare context, the token score is below
accuracy in 14 of the 15 settings, and the verbalized score is above it in all
15.  When accuracy drops, it gets closer to the token score and further from
the verbalized score.  As a result, the gap moves in favor of the token score
in all 15 settings.  In the \texttt{Guess:} slot, both scores are above
accuracy, so lower accuracy moves away from both, and the gap barely changes.

The context shift itself remains large under exact match.  Over the same 15
settings, the mean $|\Delta g|$ is $.237$ under ECE and $.235$ under AUROC,
compared with $.283$ and $.210$ under the default rule.  By point estimate,
flips go from 8 to 5 under ECE and from 10 to 11 under AUROC.

SciQ and MMLU are multiple-choice datasets, but our main setup asks the model
to answer in free text.  In the multiple-choice format, the model instead sees
the options and answers with a letter, and the bare context already asks for a
letter only, like the answer-only context (\S\ref{sec:res-ledger}).  In this format, the size of the ECE
shift depends on the model.
It stays large for Qwen3.5-9B ($.148$ and $.343$) but is $.057$ or less for
the other models.  Under AUROC, the shift averages $.043$, and the better
signal never flips (Appendix~\ref{app:correctness}).

\subsection{Comparing the Axes}
\label{sec:res-axes}

The context moves the gap more than any other axis, and the estimator moves it
least (Table~\ref{tab:axis-size}).  Under ECE, with the token score read in the
bare context, the order is the context ($.287$), the correctness rule
($.181$), token aggregation ($.054$), and the estimator ($.027$ or less;
Appendix~\ref{app:robustness}).  The
order changes with where the token score is read.  In the \texttt{Guess:} slot,
the correctness rule moves the ECE gap by only $.005$, less than token
aggregation does ($.060$; \S\ref{sec:res-correctness}).

\begin{table}[t]
\footnotesize
\centering
\setlength{\tabcolsep}{2.5pt}
\renewcommand{\arraystretch}{0.95}
\resizebox{\columnwidth}{!}{%
\begin{tabular}{@{}lcccc@{}}
\toprule
 & \multicolumn{2}{c}{ECE} & \multicolumn{2}{c}{AUROC} \\
\cmidrule(lr){2-3}\cmidrule(lr){4-5}
Axis, token score read in & bare & \texttt{Guess:} & bare & \texttt{Guess:} \\
\midrule
Context
  & \multicolumn{2}{c}{$.287$} & \multicolumn{2}{c}{$.189$} \\
Aggregation
  & $.054$ & $.060$ & $.070$ & $.026$ \\
Correctness rule$^{*}$
  & $.181$ & $.005$ & $.033$ & $.027$ \\
Estimator
  & $.005$--$.027$ & $.004$--$.010$ & -- & -- \\
\bottomrule
\end{tabular}
}
\caption{Mean $|\Delta g|$ for each axis, with the token score read in the
bare context or the \texttt{Guess:} slot.  $^{*}$Exact match versus the default,
over the 15 settings without TruthfulQA.  The estimators are alternatives for
ECE; AUROC has no estimator choice.}
\label{tab:axis-size}
\end{table}

\begin{table*}[t]
\footnotesize
\centering
\setlength{\tabcolsep}{3pt}
\begin{tabular}{@{}>{\raggedright\arraybackslash}p{0.17\textwidth}>{\raggedright\arraybackslash}p{0.6\textwidth}>{\raggedright\arraybackslash}p{0.17\textwidth}@{}}
\toprule
Choice & What to report & Evidence \\
\midrule
Generating prompt & the prompt and decoding that produced the answer &
  \S\ref{sec:res-ledger} \\
Scored answer & that the token score and the label use the model's answer &
  App.~\ref{app:goldstring} \\
Scoring context & the full prompt, with the chat template, under which the
  token score is read, and where the answer starts &
  \S\ref{sec:res-context}, \S\ref{sec:res-ledger} \\
Token aggregation & how token probabilities are combined &
  \S\ref{sec:res-axes} \\
Correctness rule & the rule and the answer format & \S\ref{sec:res-correctness} \\
Metrics & a ranking metric such as AUROC beside ECE &
  \S\ref{sec:res-context} \\
Estimator & the ECE estimator and its bins & App.~\ref{app:robustness} \\
Uncertainty & paired intervals for the gap; name a better signal only where
  both intervals exclude zero & \S\ref{sec:res-context},
  \S\ref{sec:res-correctness} \\
\midrule
\multicolumn{3}{@{}l}{\emph{Probe of verbalized confidence}} \\
Supplied answer & if the model rates a given answer, its source and exact
  string & \S\ref{sec:res-verbal} \\
\bottomrule
\end{tabular}
\caption{Reporting items for a verbalized-versus-token comparison.}
\label{tab:checklist}
\end{table*}

Even two axes together allow a wide range of gaps.  Across the six protocols
that combine the three contexts with the two aggregations, the largest point
estimate of $g$ exceeds the smallest by $.336$ under ECE and $.230$ under AUROC
on average (Appendix~\ref{app:fullgrid}).  Both ranges are larger than the
bare-context gap itself ($.304$ and $.145$).

\subsection{Supplied Answers}
\label{sec:res-verbal}

The main comparison rates the model's own answer.  Here we fix the rated answer
ourselves and look at what the stated confidence responds to
(\S\ref{sec:supplied}).  These results concern the level of verbalized
confidence, not the gap.

Stated confidence changes with how a correct answer is written.  On TriviaQA,
supplying the canonical reference instead of the first listed alias raises
confidence by $.114$ ($[.105,.123]$), although both are correct.  The rise is
limited to items where the two strings differ ($.152$); where they are the
same, the difference is $-.002$.  The canonical form is also closer to how the
models answer: it matches the model's own correct answer in $77\%$ of cases,
against $18\%$ for the first alias.

Stated confidence separates a right answer from a plausible wrong one only
weakly.  The canonical reference gets $.056$ ($[.050,.062]$) more confidence
than another model's wrong answer to the same question.  It gets less in $22\%$
of pairs and the same in $46\%$.  When the plausible wrong
answer is also a wrong option listed by the benchmark, the margin is $.083$
($[.045,.121]$).  As a sanity check, off-type
answers get $.183$ on average.

On questions the model got wrong, stated confidence leans toward the model's
own answer.  Its wrong answer gets $.044$ ($[.037,.050]$) more confidence than
another model's wrong answer to the same question, and about as much as the
canonical reference (a difference of $.008$).  When the model's answer is
correct but written differently from the canonical reference, the two get about
the same confidence ($.005$, $[-.002,.012]$).  The lean resembles the
self-preference of LLM evaluators, which score their own outputs higher than
others' \citep{Panickssery2024}.  Our model is not told who wrote a supplied
answer, so the lean may reflect agreement with its own belief rather than
authorship, much as \citet{Wataoka2024} link self-preference to how familiar
a text is to the model (Appendix~\ref{app:provenance}).

\begin{takeaway}
For a fixed answer, stated confidence depends on how the answer is written,
separates right from wrong answers only weakly, and leans toward the model's
own wrong answers.
\end{takeaway}

\subsection{Implications for Reporting}
\label{sec:discussion}

The gap between verbalized and token confidence depends on the protocol as
well as on the two signals.  Table~\ref{tab:checklist} turns our results into
reporting items.  Each item changed the gap, the verbalized score, or the
number of flips in our experiments, or is a rule whose violation did.  ECE
compares a score with accuracy, so its verdict depends on accuracy, while
AUROC does not (\S\ref{sec:res-context}); a report should therefore give both.

No context is a neutral token baseline, so the choice should follow the use
of the score and be reported.  If a model scores its own answer to decide
whether to give it, the token score should be read in the prompt that produced
the answer.  That is the context in use, and there the token score ranks the
answers better than in the bare context (Table~\ref{tab:main};
Appendix~\ref{app:reciprocal}).  If the score checks an answer from elsewhere,
its token score is lower than for an answer generated in the same context
(\S\ref{sec:res-ledger}), so a threshold set on the model's own answers does
not carry over.  The verbalized score also separates right from wrong supplied
answers only weakly and depends on their wording (\S\ref{sec:res-verbal}).
Such a check should be calibrated on answers from the same source.

Confidence used for abstention needs the same care.  The token score changes
level between contexts, from $.149$ to $.732$ on average
(\S\ref{sec:res-context}), and a calibration map fitted in one context does not
carry over to the other (Appendix~\ref{app:robustness}).  A threshold should
be set in the context where the score is used.


\section{Conclusion}
\label{sec:conclusion}

LLM confidence is sensitive to the protocol used to measure it.  Even with the
answer and its label fixed, the context in which the token score is read
shifts the gap between verbalized and token confidence by about its own size
and changes how the token score ranks the answers.  Within that context, the
instruction matters more than where the answer begins.  Other choices, such as
the correctness rule and token aggregation, can also change which score looks
better.  Stated confidence also depends on how the rated answer is written.
Reports of LLM confidence and comparisons of calibration should therefore fix
and report the measurement protocol first.  Table~\ref{tab:checklist} lists
what to report.  In practice, confidence scores should be checked under the
prompt in which they are used.

\section*{Limitations}
\label{sec:limitations}

Our experiments cover English short-answer QA with open 7--9B models, with
larger and base models as checks.  Closed models, long-form answers, multi-turn
dialogue, and reasoning before the answer remain to be tested.  We use greedy
decoding, one verbalized prompt with a 0--100 variant, and string-matching
correctness rather than human grading.

Our design scores a fixed answer in contexts that did not all produce it.  It
isolates the score of one answer rather than comparing two prompting systems
end to end, and the transfer check with answer-only answers tests whether the
shift carries over, not a causal effect of where an answer comes from.  The
prompt ablation reports changes from text edits along stated paths, so we do
not attribute the shift to any single prompt element.  The review of published
comparisons includes only studies that compare both signals and state a
conclusion; it shows that conventions differ, not how common each is.  Finally,
intervals are not corrected for multiple comparisons, and we report flips,
which depend on intervals near zero, after the size of the shift.

\section*{Ethical Considerations}
\label{sec:ethics}

We study released open models on public QA benchmarks.  We do not train or
deploy models, collect user data, or involve human participants, and we release
our code, prompts, labels, and model outputs.  The benchmarks may contain
outdated facts or offensive content; we use them only for aggregate evaluation.

The main risk this work points to is misplaced trust in a confidence score.
Systems that abstain, escalate, or verify answers based on confidence inherit
the protocol under which that confidence was measured, and a threshold or
calibration set under one prompt may not hold under another
(\S\ref{sec:discussion}).  A favorable comparison reported without its
protocol could encourage such misuse in high-stakes settings.  We therefore
recommend reporting the protocol and validating confidence under the deployed
prompt, rather than treating either score as a general measure of reliability.

\bibliography{refs/refs}

\clearpage
\appendix

\section{Reproducibility Details}
\label{app:repro}

\paragraph{Models.}  Table~\ref{tab:app-models} lists the checkpoints.  The
main analysis uses the instruction-tuned checkpoints.  We also repeat it on
the base checkpoints and on four larger instruction-tuned models.

\begin{table*}[t]
\footnotesize
\centering
\setlength{\tabcolsep}{4pt}
\resizebox{\textwidth}{!}{%
\begin{tabular}{@{}llllll@{}}
\toprule
Model & Developer & Instruction-tuned checkpoint & Revision & Base checkpoint & Revision \\
\midrule
Llama-3-8B \citep{Llama3} & Meta &
  \texttt{Meta-Llama-3-8B-Instruct} & \texttt{8afb486} & \texttt{Meta-Llama-3-8B} & \texttt{8cde5ca} \\
Mistral-7B-v0.3 \citep{Mistral7B} & Mistral AI &
  \texttt{Mistral-7B-Instruct-v0.3} & \texttt{c170c70} & \texttt{Mistral-7B-v0.3} & \texttt{caa1feb} \\
Qwen2.5-7B \citep{Qwen25} & Alibaba (Qwen) &
  \texttt{Qwen2.5-7B-Instruct} & \texttt{a09a354} & \texttt{Qwen2.5-7B} & \texttt{d149729} \\
Qwen3.5-9B \citep{Qwen35} & Alibaba (Qwen) &
  \texttt{Qwen3.5-9B} & \texttt{c202236} & \texttt{Qwen3.5-9B-Base} & \texttt{68c46c4} \\
Ministral-3-8B \citep{Ministral3} & Mistral AI &
  \texttt{Ministral-3-8B-Instruct-2512-BF16} & \texttt{f6fae97} &
  \texttt{Ministral-3-8B-Base-2512} & \texttt{d4883f9} \\
\midrule
Qwen2.5-14B \citep{Qwen25} & Alibaba (Qwen) &
  \texttt{Qwen2.5-14B-Instruct} & \texttt{cf98f3b} & -- & -- \\
Qwen2.5-32B \citep{Qwen25} & Alibaba (Qwen) &
  \texttt{Qwen2.5-32B-Instruct} & \texttt{5ede1c9} & -- & -- \\
Qwen2.5-72B \citep{Qwen25} & Alibaba (Qwen) &
  \texttt{Qwen2.5-72B-Instruct} & \texttt{495f393} & -- & -- \\
Ministral-3-14B \citep{Ministral3} & Mistral AI &
  \texttt{Ministral-3-14B-Instruct-2512-BF16} & \texttt{3cea74c} & -- & -- \\
\bottomrule
\end{tabular}
}
\caption{Models.  The last four rows are the larger models, which have no base checkpoint in our runs.  Checkpoint names are Hugging Face repository names without
the organization prefix; revisions are the first seven characters of the
commit.  Qwen3.5-9B runs without thinking mode.}
\label{tab:app-models}
\end{table*}

\paragraph{Data.}  Table~\ref{tab:app-data} lists the data sources, and
Table~\ref{tab:app-n} the number of items per setting after we drop unparsed
rows.

\paragraph{Licenses and content.}  SciQ is released under CC BY-NC 3.0,
TruthfulQA under Apache-2.0, MMLU under MIT, and the TriviaQA repository under
Apache-2.0.  We use the models under their released licenses: Apache-2.0 for
all except Llama-3-8B (Llama 3 Community License) and Qwen2.5-72B (Qwen
License).  All are used for research evaluation, which these terms allow.  Our
released outputs contain the benchmark questions, so they inherit these terms;
in particular, the SciQ records are for non-commercial use only.  The
benchmarks are public and contain no data about private individuals, and we add
none.  We do not filter questions or outputs for content.  The model outputs, including
the supplied wrong answers, may contain false or objectionable statements; the
release documents this and is intended only for reproducing our measurements.

\begin{table*}[t]
\footnotesize
\centering
\setlength{\tabcolsep}{4pt}
\resizebox{\textwidth}{!}{%
\begin{tabular}{@{}llllr@{}}
\toprule
Dataset & Source split & Selection & Gold answers & Items \\
\midrule
TriviaQA \citep{Joshi2017} & \texttt{rc.nocontext}, validation &
  first 1,000 questions & aliases and value & 1,000 \\
SciQ \citep{Welbl2017} & test & first 1,000 questions & correct answer & 1,000 \\
TruthfulQA \citep{Lin2022} & \texttt{generation}, validation &
  all questions & best and correct answers & 817 \\
MMLU \citep{Hendrycks2021} & test &
  four subjects$^{\ddagger}$, first 500 per subject & text of the correct choice & 1,155 \\
\bottomrule
\end{tabular}
}
\caption{Data.  $^{\ddagger}$High-school physics, philosophy, high-school
government and politics, and miscellaneous.  The first three subjects have
fewer than 500 test questions, so we use all of them.  SciQ and MMLU are
asked as open questions, without their options, except in the
multiple-choice format (Appendix~\ref{app:correctness}).}
\label{tab:app-data}
\end{table*}

\begin{table}[t]
\footnotesize
\centering
\setlength{\tabcolsep}{4pt}
\begin{tabular}{@{}lrrrr@{}}
\toprule
Model & TriviaQA & SciQ & TruthfulQA & MMLU \\
\midrule
Llama-3-8B & 1000 & 1000 & 817 & 1155 \\
Mistral-7B-v0.3 & 997 & 992 & 789 & 1119 \\
Qwen2.5-7B & 1000 & 1000 & 817 & 1155 \\
Qwen3.5-9B & 1000 & 998 & 786 & 1142 \\
Ministral-3-8B & 999 & 984 & 734 & 1120 \\
\bottomrule
\end{tabular}
\caption{Items per setting after dropping unparsed rows.}
\label{tab:app-n}
\end{table}

\paragraph{Generation.}  Answers use greedy decoding with at most 32 new
tokens.  We keep each checkpoint's released generation configuration with
sampling turned off; for the Qwen2.5 models, this keeps a repetition penalty
of 1.05.  Main answers are generated one question at a time.  The answer-only
answers are generated in batches of 16 with left padding for Llama-3-8B,
Mistral-7B-v0.3, and Qwen2.5-7B, and one at a time for Qwen3.5-9B and
Ministral-3-8B; the multiple-choice answers follow the same rule.  Token scores are computed in batches of 16, except for
Qwen3.5-9B, whose scores change with padding, and for Ministral-3-8B in the
multiple-choice format and on the answer-only answers; these we score one
input at a time.  Models run in float16, except Qwen2.5-72B, which runs in bfloat16.

\paragraph{Chat templates.}  We wrap each prompt as a user turn with the
model's released chat template and let the template open the assistant turn.
The Llama-3 and Mistral-7B templates add no system prompt.  The Qwen2.5
templates add the default system prompt ``You are Qwen, created by Alibaba
Cloud.\ You are a helpful assistant.'', and the Ministral-3 templates add their
long default system prompt, whose date placeholder stays unfilled; we keep
both.  Qwen3.5-9B runs with thinking turned off, so its assistant turn opens
with an empty thinking block before the marker or the answer.  We load the
Ministral-3 tokenizer as released, without the optional fix to its
pre-tokenization pattern; the fix does not change the tokens of our inputs.

\paragraph{Prompts and contexts.}  Table~\ref{tab:app-prompts} gives the
prompts.  In the bare context, the user turn is the bare prompt and the answer
starts the assistant turn.  In the \texttt{Answer:} slot and the answer-only
context, the final \texttt{Answer:} of the bare prompt or of the answer-only
instruction moves from the user turn to the start of the assistant turn.  In the
\texttt{Guess:} slot, the user turn is the confidence prompt and the assistant
turn starts with \texttt{Guess:}.  In the bare context, the answer starts the
assistant turn without a leading space.  After \texttt{Answer:} or
\texttt{Guess:}, it follows a single space, so the first token of the answer can
differ between the bare context and the \texttt{Answer:} slot.  We tokenize
the context and the parsed answer string separately and join the tokens, so
the token score is computed on the parsed answer, not on the generated
tokens.  The answer-only answers
(\S\ref{sec:contexts}) are generated with the full answer-only instruction as
the user turn.

\begin{table*}[t]
\footnotesize
\centering
\begin{tabular}{@{}p{0.17\textwidth}p{0.78\textwidth}@{}}
\toprule
Prompt & Text \\
\midrule
Confidence prompt \citep{Tian2023} &
\texttt{Provide your best guess and the probability that it is correct (0.0 to
1.0) for the following question. Give ONLY the guess and probability, no other
words or explanation. For example:}\newline\newline
\texttt{Guess: <most likely guess, as short as possible; not a complete
sentence, just the guess!>}\newline
\texttt{Probability: <the probability between 0.0 and 1.0 that your guess is
correct, without any extra commentary whatsoever; just the
probability!>}\newline\newline
\texttt{The question is: \{question\}} \\
\midrule
Confidence prompt, 0--100 scale (Appendix~\ref{app:promptscale}) &
\texttt{Provide your best guess and the probability that it is correct (0 to 100
as an integer percentage) for the following question. Give ONLY the guess and
probability, no other words or explanation.}\newline\newline
\texttt{Guess: <short answer>}\newline
\texttt{Probability: <integer 0 to 100>}\newline\newline
\texttt{The question is: \{question\}} \\
\midrule
Multiple-choice confidence prompt (Appendix~\ref{app:correctness}) &
\texttt{Choose the best answer to the following multiple-choice question. Give
ONLY the option letter and the probability that it is correct (0.0 to 1.0), with
no explanation. Use exactly this format:}\newline\newline
\texttt{Guess: <A, B, C, or D>}\newline
\texttt{Probability: <number from 0.0 to 1.0>}\newline\newline
\texttt{Question: \{question\}}\newline\texttt{A. \{option\}} \ldots\
\texttt{D. \{option\}} \\
\midrule
Multiple-choice letter-only prompt &
\texttt{Choose the best answer to the following multiple-choice question. Give
ONLY one option letter (A, B, C, or D), with no explanation.}\newline\newline
\texttt{Question: \{question\}}\newline\texttt{A. \{option\}} \ldots\
\texttt{D. \{option\}} \\
\midrule
Bare prompt &
\texttt{Question: \{question\}}\newline\texttt{Answer:} \\
\midrule
Answer-only instruction &
\texttt{Provide your best guess for the following question. Give ONLY the
guess, no other words or explanation. Keep the guess as short as possible; do
not write a complete sentence.}\newline\newline
\texttt{The question is: \{question\}}\newline\texttt{Answer:} \\
\bottomrule
\end{tabular}
\caption{Prompts.  \texttt{\{question\}} is replaced by the question.  We
show the line breaks of the prompts; other long lines only wrap.}
\label{tab:app-prompts}
\end{table*}

\paragraph{Parsing.}  The answer is the first non-empty text after a
\texttt{Guess:} label; without that label, it is the first line that is not a
label.  The verbalized score is the number after a \texttt{Probability:} label;
if no label matches, the parser takes the first standalone number between 0
and 1 in the output.  We drop rows where either is missing.  The fallback sets
the score of 331 of the 19,604 scored rows, mostly for Ministral-3-8B, which
often writes the probability in bold, and Llama-3-8B, which sometimes leaves
out the labels.  Dropping these rows leaves the main result unchanged: the
mean $|\Delta g|$ is $.288$ under ECE and $.188$ under AUROC, with 8 and 13
flips by point estimate.  We do not strip Markdown from answers, so 342 of the
3,837 scored answers of Ministral-3-8B keep their asterisks.  The lowest parse
rate, $.898$, is for Ministral-3-8B on TruthfulQA, where long answers often
reach the 32-token limit before the probability line.

\paragraph{Metrics.}  ECE sorts the $n$ answers into 10 equal-width confidence
bins and sums, over bins $b$, $\frac{n_b}{n}\,|\mathrm{acc}_b -
\mathrm{conf}_b|$, where $n_b$ is the bin size and $\mathrm{acc}_b$ and
$\mathrm{conf}_b$ are the mean correctness and the mean confidence in the bin.
The bins are $[k/10,(k+1)/10)$, and the last one also includes 1.
AUROC is the probability that a randomly chosen correct answer receives higher
confidence than a randomly chosen wrong one; ties count one half.  For AURC, we keep the $k$ most
confident answers for each $k=1,\dots,n$, compute the error rate among them, and
average over $k$.  Answers with tied confidence are ordered at random, and we use
the expected value over that ordering.

\paragraph{Normalization.}  Normalization lowercases the text, replaces every
character other than an ASCII letter, digit, or whitespace with a space, and
merges repeated whitespace.

\paragraph{Intervals.}  All intervals are 95\% percentile intervals from a
paired bootstrap with 10,000 resamples.  Within a setting, we resample answers
and recompute every score on the same resample.  Each setting has its own
random seed, fixed by its model and dataset.  For a mean over settings, we
average the resampled values across settings and take the percentiles of that
average.

\paragraph{Environment.}  We run on NVIDIA A100 80GB GPUs with PyTorch 2.5.1
and Transformers 5.8.1.  The checkpoint revisions are the Hugging Face commits
listed in Table~\ref{tab:app-models}.

\section{Further Checks of the Context Shift}
\label{app:checks}

Table~\ref{tab:app-checks} repeats the main comparison under three changes: a
0--100 confidence scale, larger models, and base checkpoints.  Each row uses
the same metrics and intervals as Table~\ref{tab:main}, and the same contexts
except where Appendix~\ref{app:base} notes otherwise.

\begin{table*}[t]
\footnotesize
\centering
\setlength{\tabcolsep}{4pt}
\begin{tabular}{@{}lrcccccc@{}}
\toprule
 & & \multicolumn{2}{c}{mean $|\Delta g|$} &
   \multicolumn{2}{c}{interval excludes 0} & AUROC shift & flips \\
\cmidrule(lr){3-4}\cmidrule(lr){5-6}
Check & Settings & ECE & AUROC & ECE & AUROC & same direction & ECE; AUROC \\
\midrule
Main comparison (Table~\ref{tab:main}) & 20 & $.287$ & $.189$ & 19 & 17 & 20 & 8 (8); 8 (13) \\
\midrule
0--100 scale & 20 & $.274$ & $.210$ & 18 & 18 & 20 & 6 (8); 6 (12) \\
\midrule
Qwen2.5-14B & 4 & $.354$ & $.269$ & 4 & 4 & 4 & 2 (2); 1 (2) \\
Qwen2.5-32B & 4 & $.359$ & $.295$ & 4 & 4 & 4 & 2 (2); 2 (3) \\
Qwen2.5-72B$^{\dagger}$ & 4 & $.337$ & $.288$ & 4 & 4 & 4 & 1 (2); 2 (4) \\
Ministral-3-14B & 4 & $.306$ & $.202$ & 4 & 4 & 4 & 2 (2); 0 (3) \\
\midrule
Base checkpoints & 20 & $.140$ & $.137$ & 11 & 12 & 9 & 4 (6); 4 (6) \\
\bottomrule
\end{tabular}
\caption{The context shift (bare context $\to$ \texttt{Guess:} slot) under
three checks.  ``Same direction'': settings where token AUROC is higher in the
\texttt{Guess:} slot.  Flips: fully resolved (point estimate).
$^{\dagger}$TriviaQA, SciQ, and TruthfulQA use their first 250 questions.}
\label{tab:app-checks}
\end{table*}

\subsection{Integer 0--100 Elicitation}
\label{app:promptscale}

Here the confidence prompt asks for an integer percentage from 0 to 100
(Table~\ref{tab:app-prompts}), and the \texttt{Guess:} slot uses this prompt.
We keep only outputs that contain the \texttt{Probability:} label, which leaves
16,328 of the 19,320 parsed rows ($.845$).  Llama-3-8B often leaves out the
label, so only $.259$ of its rows remain; each other model keeps at least
$.991$.  The shift stays about as large as in the main comparison, and token
AUROC again rises in the \texttt{Guess:} slot in all 20 settings.

\subsection{Larger Models}
\label{app:larger}

We run three larger Qwen2.5 models and Ministral-3-14B on the four datasets.
To limit compute, Qwen2.5-72B answers only the first 250 questions of
TriviaQA, SciQ, and TruthfulQA.  In every setting of every larger model, the
interval of $\Delta g$ excludes zero under both metrics, and token AUROC rises
in the \texttt{Guess:} slot.  The mean $|\Delta g|$ is at least as large as in
the main comparison.

\subsection{Base Checkpoints}
\label{app:base}

Base checkpoints often ignore the requested format, so many rows drop out.
Of the 3,972 questions per model, we keep $.158$ for Llama-3-8B, $.369$ for
Mistral-7B-v0.3, $.930$ for Qwen2.5-7B, $.971$ for Qwen3.5-9B, and $.119$ for
Ministral-3-8B.  The tokenizers of the two Qwen base models have a chat
template, and we use it.  The other three have none, so the prompt and the
answer form one plain text, and the \texttt{Answer:} slot is the same as the
bare context.  For these three models, the \texttt{Guess:} slot is the
confidence prompt followed by a new line and \texttt{Guess:}, the answer in
the bare context also follows a space, and answers are generated without the
tokenizer's beginning-of-sequence token but scored with it.  The shift is smaller in base checkpoints, and its direction
under AUROC is mixed: token AUROC rises in the \texttt{Guess:} slot in 9 of
the 20 settings.

\section{Prompt Ablation}
\label{app:ablation}

\paragraph{Design.}  The \texttt{Answer:} slot and the \texttt{Guess:} slot
differ in both the user prompt and the marker.  We cross four user prompts with
the two markers \texttt{Answer:} and \texttt{Guess:} at the start of the
assistant turn, and score each model's main answers in all eight cells.  The
four prompts are the question alone (the bare prompt without its final
\texttt{Answer:}), the answer-only instruction (also without its final
\texttt{Answer:}), the confidence prompt without its probability request, and
the full confidence prompt (Table~\ref{tab:app-prompts}).  To remove the
probability request, we delete its three mentions: ``and the probability that
it is correct (0.0 to 1.0)'', ``and probability'', and the
\texttt{Probability:} line of the example.  Three cells are contexts of the
main text: (question, \texttt{Answer:}) is the \texttt{Answer:} slot,
(answer-only instruction, \texttt{Answer:}) is the answer-only context, and
(confidence prompt, \texttt{Guess:}) is the \texttt{Guess:} slot.  Re-scoring
the \texttt{Answer:} slot and the \texttt{Guess:} slot for the main answers,
and the answer-only context for the answer-only answers, reproduces the stored
token scores exactly.

\begin{table}[t]
\footnotesize
\centering
\setlength{\tabcolsep}{3pt}
\resizebox{\columnwidth}{!}{%
\begin{tabular}{@{}lcccc@{}}
\toprule
 & \multicolumn{2}{c}{token AUROC} & \multicolumn{2}{c}{mean token score} \\
\cmidrule(lr){2-3}\cmidrule(lr){4-5}
User prompt & \texttt{Answer:} & \texttt{Guess:} & \texttt{Answer:} & \texttt{Guess:} \\
\midrule
Question alone & $.549$ & $.611$ & $.185$ & $.229$ \\
Answer-only instruction & $.722$ & $.719$ & $.562$ & $.582$ \\
Confidence prompt without probability & $.742$ & $.759$ & $.608$ & $.663$ \\
Confidence prompt & $.709$ & $.757$ & $.612$ & $.732$ \\
\bottomrule
\end{tabular}
}
\caption{Prompt ablation: the same answers scored in eight cells, mean over 20
settings.  The verbalized AUROC is $.702$ in every cell.}
\label{tab:app-ablation}
\end{table}

\paragraph{Results.}  Table~\ref{tab:app-ablation} shows where token AUROC
changes.  Replacing the question with the answer-only instruction raises it in
both columns.  The marker matters with the question alone ($.549$ to $.611$)
but hardly with the answer-only instruction ($.722$ to $.719$).  The full step,
from the \texttt{Answer:} slot to the \texttt{Guess:} slot, raises token
AUROC by $.208$ ($[.198,.218]$).  Along the \texttt{Guess:} path, switching
the marker under the question alone adds $.062$ ($[.056,.067]$), the
answer-only instruction $.109$ ($[.100,.117]$), the rest of the output
template $.040$ ($[.035,.045]$), and the probability request $-.002$
($[-.007,.004]$).  Swapping the marker under the answer-only instruction
changes token AUROC by $-.003$ ($[-.005,.000]$).  The instruction effect is
positive for every model, from $.060$ (Mistral-7B-v0.3) to $.225$
(Qwen2.5-7B).

\paragraph{Order of the edits.}  The size of each edit depends on the path.
With the \texttt{Answer:} marker fixed, as in Table~\ref{tab:ledger}, the
answer-only instruction adds $.173$, the output template $.020$, the
probability request $-.033$, and the final switch to \texttt{Guess:} $.048$.
We therefore describe each number as the change from one edit along a stated
path, not as the share of the shift caused by one part of the prompt.

\paragraph{ECE.}  The same edits move the ECE gap by different amounts.  On the
\texttt{Guess:} path, the mean $|\Delta g|$ is $.070$ for the marker under
the question alone, $.181$ for the answer-only
instruction, $.053$ for the output template, $.060$ for the probability
request, and $.020$ for the marker under the answer-only instruction, against
$.266$ for the full step from the \texttt{Answer:} slot to the \texttt{Guess:} slot.  On the \texttt{Answer:} path, the probability
request moves it by $.039$, the answer-only instruction by $.175$, and the
move from the bare context to the \texttt{Answer:} slot by $.031$; these are
the values in \S\ref{sec:res-ledger}.  So the probability request leaves the ranking
almost unchanged but does shift the level of the token score.

\paragraph{Answers from the answer-only instruction.}  We repeat the grid with
the answer-only answers (\S\ref{sec:contexts}).  The pattern is similar: the
answer-only instruction raises token AUROC by $.135$ ($[.127,.144]$), the
probability request changes it by $-.002$ ($[-.007,.003]$), and the marker
under the answer-only instruction by $-.011$ ($[-.014,-.007]$).  These answers
were generated under the answer-only instruction, so for them this edit also
moves the context toward the prompt that produced them.

\paragraph{Selection.}  Each answer gets a higher token score in the context
where greedy decoding chose it.  We take the questions where the main answer
and the answer-only answer differ ($.595$ of matched questions) and compare
their mean token scores.  The main answer scores $.121$ lower in the
answer-only context and $.318$ higher in the \texttt{Guess:} slot, where it was
produced.  The answer-only context uses the instruction that produced the
answer-only answer but differs from its generating context in where
\texttt{Answer:} falls.  The difference
between the two contexts, $.438$ ($[.431,.445]$), ranges from $.293$ to $.611$
across models.

\section{Answers from Another Prompt}
\label{app:prompt-slot}

In the bare context and the \texttt{Answer:} slot, the token score describes
an answer that the model generated under another prompt.  To see whether this
matters, we use two answers from the same model to each question: the main
answer $\hat a_i$ from the confidence prompt and the answer-only answer $\hat
b_i$ from the answer-only instruction (Table~\ref{tab:app-prompts}).  We read
the token score of both answers in the bare context and in the \texttt{Guess:}
slot.  An average setting has 980 questions with both answers, and the two
answers differ on $.595$ of them (exact string match).  When the two strings
are the same, their token scores differ by at most $.010$; this comes from
batched scoring, and Qwen3.5-9B, scored one input at a time, shows no
difference.

\begin{table}[t]
\footnotesize
\centering
\setlength{\tabcolsep}{3pt}
\resizebox{\columnwidth}{!}{%
\begin{tabular}{@{}lcccc@{}}
\toprule
 & \multicolumn{2}{c}{mean token score} & \multicolumn{2}{c}{$\hat a_i$ minus $\hat b_i$} \\
\cmidrule(lr){2-3}\cmidrule(lr){4-5}
Context & $\hat a_i$ & $\hat b_i$ & all questions & different strings \\
\midrule
Bare context & $.149$ & $.157$ & $-.008$ \ci{[-.010,-.006]} & $-.016$ \ci{[-.019,-.012]} \\
\texttt{Guess:} slot & $.732$ & $.545$ & $+.187$ \ci{[+.183,+.191]} & $+.318$ \ci{[+.312,+.323]} \\
\bottomrule
\end{tabular}
}
\caption{Token scores of the main answer $\hat a_i$ (from the confidence
prompt) and the answer-only answer $\hat b_i$ for the same questions, mean over
20 settings, with 95\% paired bootstrap intervals.}
\label{tab:app-prompt-slot}
\end{table}

Table~\ref{tab:app-prompt-slot} shows the result.  In the bare context, the
main answer scores slightly lower than the answer-only answer ($-.008$).  In
the \texttt{Guess:} slot, where the main answer was generated, it scores
$.187$ higher.  The difference between the two contexts is $.195$
($[.191,.199]$) and ranges from $.143$ to $.255$ across models.  So the bare context scores the two answers alike, although neither was
generated there, while the \texttt{Guess:} slot favors the answer generated
there.

\subsection{Transfer Check}
\label{app:reciprocal}

We also repeat the main comparison with the answer-only answers in place of the
main answers.  Each answer-only answer keeps its own correctness label.  Its
verbalized score comes from supplying it after \texttt{Guess:}, with the same
format as in \S\ref{sec:supplied}, because the answer-only instruction does not ask for a
probability.  We read its token score in three contexts: the bare context, the
prompt that generated it, and the \texttt{Guess:} slot.

The shift from the bare context to the \texttt{Guess:} slot remains.  Its mean
$|\Delta g|$ is $.181$ under ECE and $.161$ under AUROC, and its interval
excludes zero in 17 and 15 of 20 settings.  The better signal flips in 6
settings under ECE and 4 under AUROC (7 and 10 by point estimate).  For these
answers, token AUROC is highest in the prompt that generated them ($.775$),
against $.730$ in the \texttt{Guess:} slot and $.573$ in the bare context; the
verbalized AUROC is $.708$.  Because the verbalized score here rates a supplied
answer, this check tests whether the shift carries over to other answers, not
the main comparison itself.

\section{Scoring the Reference String}
\label{app:goldstring}

A token baseline breaks the alignment rule (\S\ref{sec:events}) if it scores
the reference answer instead of the model's answer.  To measure what this
changes, we read the token score of the canonical reference
(Appendix~\ref{app:provenance}) in the three contexts and keep the verbalized score and the label of the model's answer.
The reference is the same string as the model's answer, ignoring case, for
$.265$ of the answers.

\begin{table}[t]
\footnotesize
\centering
\setlength{\tabcolsep}{3pt}
\resizebox{\columnwidth}{!}{%
\begin{tabular}{@{}lcccccc@{}}
\toprule
 & \multicolumn{2}{c}{mean token score} & \multicolumn{2}{c}{token AUROC} &
   \multicolumn{2}{c}{change in $g$} \\
\cmidrule(lr){2-3}\cmidrule(lr){4-5}\cmidrule(lr){6-7}
Context & answer & reference & answer & reference & ECE & AUROC \\
\midrule
Bare context & $.149$ & $.101$ & $.568$ & $.617$ & $.054$ (2) & $.061$ (4) \\
\texttt{Answer:} slot & $.185$ & $.127$ & $.549$ & $.611$ & $.064$ (3) & $.074$ (4) \\
\texttt{Guess:} slot & $.732$ & $.273$ & $.757$ & $.783$ & $.224$ (2) & $.093$ (7) \\
\bottomrule
\end{tabular}
}
\caption{Token score of the model's answer and of the reference answer, mean
over 20 settings.  Change in $g$: mean $|\Delta g|$ when the reference replaces
the answer, with the number of settings where the better signal flips (point
estimate) in parentheses.}
\label{tab:app-goldstring}
\end{table}

Table~\ref{tab:app-goldstring} shows two effects.  First, the reference gets a
lower token score than the model's answer in every context, and by far the
most in the \texttt{Guess:} slot ($.732$ to $.273$), where the model generated
its answer.  So under ECE, the substitution moves the gap most in that slot
($.224$, against $.054$ in the bare context).  This ECE compares a score for
one string with a label for another, so it is not a calibration result for
either answer.

Second, the reference often ranks the model's answers better than the answers
themselves do.  Mean token AUROC is higher in every context, and it is higher
in 16, 16, and 12 of the 20 settings in the three contexts.  The reference separates
right from wrong answers more sharply.  In the \texttt{Guess:} slot, it gets
$.460$ on average when the model's answer is correct and $.105$ when it is
wrong, while the model's own answer gets $.833$ and $.658$.  The substitution thus makes the token score look better, and
under AUROC the better signal flips in 4 to 7 of the 20 settings.

\section{Correctness Rule and Answer Format}
\label{app:correctness}

\paragraph{Exact match.}  Under exact match, the answer must equal a gold
answer after normalization (Appendix~\ref{app:repro}).  We leave TruthfulQA out
of exact-match summaries, because its exact-match accuracy is $.023$ to $.048$
across models.  Over the other 15 settings, exact match lowers accuracy from
$.493$ to $.387$ on average.  This raises the bare-context ECE gap in all 15
settings, by $.181$ on average and by at least $.039$, while the
\texttt{Guess:} gap moves by only $.005$ (\S\ref{sec:res-correctness}).  The
context shift stays large: its mean $|\Delta g|$ goes from $.283$ to $.237$
under ECE and from $.210$ to $.235$ under AUROC.  By point estimate, ECE flips
fall from 8 to 5 and AUROC flips go from 10 to 11.  The three ECE flips that
disappear are all on SciQ (Llama-3-8B, Ministral-3-8B, and Qwen2.5-7B), where
the bare gap was negative and becomes positive, like the \texttt{Guess:} gap.

\paragraph{Multiple-choice format.}  SciQ and MMLU are multiple-choice
datasets.  In this format, the model sees four lettered options and answers
with a letter, which is correct if it matches the gold letter.  For SciQ, the
options are the correct answer and the three listed distractors in a shuffled
order; for MMLU, they are the four given choices.  This format uses its own two prompts
(Table~\ref{tab:app-prompts}).  The multiple-choice confidence prompt asks for
the option letter and its probability, and the \texttt{Guess:} slot uses it.
The letter-only prompt asks for one option letter without a probability; the
bare context uses it, and the \texttt{Answer:} slot adds \texttt{Answer:} at
the start of the assistant turn.  The model writes at most 20 new tokens, and
the SciQ options are shuffled with a fixed seed for each question.  Unlike the bare prompt of the main analysis, this prompt contains
an answer-only instruction, so it is closer to the answer-only context of
\S\ref{sec:res-ledger}.

\begin{table}[t]
\footnotesize
\centering
\setlength{\tabcolsep}{3pt}
\resizebox{\columnwidth}{!}{%
\begin{tabular}{@{}lcccccc@{}}
\toprule
 & \multicolumn{2}{c}{accuracy} & \multicolumn{2}{c}{$|\Delta g|$, ECE} &
   \multicolumn{2}{c}{$|\Delta g|$, AUROC} \\
\cmidrule(lr){2-3}\cmidrule(lr){4-5}\cmidrule(lr){6-7}
Model & SciQ & MMLU & SciQ & MMLU & SciQ & MMLU \\
\midrule
Llama-3-8B & $.907$ & $.715$ & $.002$ & $.046$ & $.007$ & $.021$ \\
Mistral-7B-v0.3 & $.822$ & $.646$ & $.020$ & $.057$ & $.019$ & $.000$ \\
Qwen2.5-7B & $.929$ & $.786$ & $.010$ & $.031$ & $.005$ & $.026$ \\
Qwen3.5-9B & $.964$ & $.857$ & $.343$ & $.148$ & $.173$ & $.122$ \\
Ministral-3-8B & $.924$ & $.766$ & $.038$ & $.012$ & $.049$ & $.004$ \\
\bottomrule
\end{tabular}
}
\caption{Multiple-choice format: accuracy and the context shift (bare context
$\to$ \texttt{Guess:} slot) per setting.}
\label{tab:app-mc}
\end{table}

Table~\ref{tab:app-mc} gives the results.  Accuracy is high, from $.646$ to
$.964$.  The ECE shift is small for four models and large for Qwen3.5-9B, so
its mean over the 10 settings is $.071$, and $.027$ without Qwen3.5-9B.  Its
interval excludes zero in 7 settings under ECE and 5 under AUROC.  Under ECE,
the better signal flips in 1 setting (5 by point estimate).  Under AUROC and
AURC, it flips in none.

\section{Estimators, Bin Counts, and Calibration Transfer}
\label{app:robustness}

\paragraph{Estimators.}  Besides 10-bin ECE, we use three estimators without
fixed bins.  \emph{KDE} is a plain kernel estimate in the spirit of \citet{Zhang2020},
not their exact estimator: it estimates accuracy at each answer's confidence by
Gaussian kernel regression (bandwidth $.05$) and averages its absolute
difference from the confidence over answers.  \emph{KS} follows
\citet{GuptaK2021}: it sorts answers by confidence and takes the largest
absolute cumulative difference between correctness and confidence, divided by
$n$.  \emph{Smoothed} follows \citet{BlasiokNakkiran2024} in smoothing both the
accuracy curve and the confidence density with a Gaussian kernel, but uses a
plain kernel with Silverman's bandwidth instead of their reflected kernel and
bandwidth rule.

\begin{table}[t]
\footnotesize
\centering
\setlength{\tabcolsep}{3pt}
\resizebox{\columnwidth}{!}{%
\begin{tabular}{@{}lcccccc@{}}
\toprule
 & \multicolumn{2}{c}{bare context} & \multicolumn{2}{c}{\texttt{Guess:} slot} &
   better signal & context \\
\cmidrule(lr){2-3}\cmidrule(lr){4-5}
Estimator & mean & max & mean & max & changes & flips \\
\midrule
10-bin ECE & -- & -- & -- & -- & -- & 8 \\
KDE & $.005$ & $.012$ & $.004$ & $.014$ & 0 & 8 \\
KS & $.015$ & $.092$ & $.009$ & $.026$ & 0 & 8 \\
Smoothed & $.027$ & $.049$ & $.010$ & $.027$ & 1 & 9 \\
\bottomrule
\end{tabular}
}
\caption{Estimators.  Mean and max: $|\Delta g|$ over 20 settings when the
estimator replaces 10-bin ECE, with the token score read in the given context.
Better signal changes: settings where the estimator, not the context, flips
the better signal.  Context flips: settings where the better signal differs
between the bare context and the \texttt{Guess:} slot, by point estimate.}
\label{tab:app-estimators}
\end{table}

Table~\ref{tab:app-estimators} gives the results.  Changing the estimator moves
the gap by $.027$ or less on average.  The largest change in a single setting
is $.092$ (KS, bare context).  Only one estimator changes the better signal in
one setting (Smoothed, \texttt{Guess:} slot, Mistral-7B-v0.3 on TruthfulQA).
The context flips stay at 8 or 9 under every estimator.

\paragraph{Bin counts.}  With 5, 10, 15, 20, 30, or 50 equal-width bins, the
better signal flips between the bare context and the \texttt{Guess:} slot in 8
settings, and between the \texttt{Answer:} slot and the \texttt{Guess:} slot in
7 settings, at every bin count (point estimates).

\paragraph{Calibration transfer.}  Recalibrating the token score within
one context removes any monotone difference in its scale.  The recalibrated
scores still differ between contexts.  With isotonic regression fitted and
evaluated in the same context, the Brier score is $.196$ in the bare context
and $.170$ in the \texttt{Guess:} slot ($.200$ and $.170$ with Platt scaling).
A calibration map also does not carry over between contexts.  We fit the map
on the token scores of one context and apply it to the token scores of the
same answers in the other context, cross-fitting with 5 folds and repeating
with 10 random splits.  The penalty is the Brier score of the transferred map
minus that of a map fitted in the same context.  It is positive in all 20
settings, for both methods and both directions, and in every repeat.  On
average it is $.124$ (isotonic) and $.128$ (Platt) when a map from the
\texttt{Guess:} slot is applied to the bare context, and $.149$ and $.049$ in
the other direction.

\section{Protocol Ranges}
\label{app:fullgrid}

A reader who reimplements a comparison without the protocol has to pick one.
To see how much the pick can matter, we take each setting, compute the gap $g$
under every protocol in a set, and record two numbers: the width (the largest
minus the smallest point estimate of $g$) and whether the protocols give both
signs.  Table~\ref{tab:app-ranges} reports the mean width over the 20 settings
and the number of settings with both signs.

\begin{table}[t]
\footnotesize
\centering
\setlength{\tabcolsep}{3pt}
\resizebox{\columnwidth}{!}{%
\begin{tabular}{@{}lccccc@{}}
\toprule
 & & \multicolumn{2}{c}{ECE} & \multicolumn{2}{c}{AUROC} \\
\cmidrule(lr){3-4}\cmidrule(lr){5-6}
Protocols & Number & width & both signs & width & both signs \\
\midrule
Context & 3 & $.292$ & 8 & $.219$ & 14 \\
Context $\times$ aggregation & 6 & $.336$ & 9 & $.230$ & 17 \\
\midrule
\multicolumn{6}{@{}l}{\emph{Breaking the alignment rule}} \\
Context $\times$ scored string & 6 & $.374$ & 10 & $.287$ & 18 \\
Context $\times$ aggregation $\times$ scored string & 12 & $.405$ & 10 & $.301$ & 18 \\
\bottomrule
\end{tabular}
}
\caption{Range of the gap across protocols.  Width: largest minus smallest
point estimate of $g$, mean over 20 settings.  Both signs: settings where the
protocols disagree on the better signal.  Scored string: the model's answer or
the reference answer (Appendix~\ref{app:goldstring}).}
\label{tab:app-ranges}
\end{table}

The range grows as we add axes.  With the three contexts and the two
aggregations, the width is $.336$ under ECE and $.230$ under AUROC, larger
than the bare-context gap itself ($.304$ and $.145$).  The sign of the
comparison differs across these six protocols in 9 settings under ECE and 17
under AUROC.  Scoring the reference string widens the range further, but such
protocols break the alignment rule, so we keep them apart.

These numbers are point estimates, and a large width can partly reflect noise.
They are also a lower bound on the range a reader faces, because we vary only
the axes in the table.

\section{Supplied Answers}
\label{app:provenance}

\paragraph{Setup.}  The user turn is the confidence prompt.  We start the
assistant turn with \texttt{Guess:}, the supplied answer, a new line, and
\texttt{Probability:}, and let the model write at most 12 more tokens.  The
canonical reference is the \texttt{answer.value} field for TriviaQA and the
first gold answer of Table~\ref{tab:app-data} for the other datasets (for
TruthfulQA, the best answer).  The plausible
wrong answer is a wrong answer that another model gave to the same question and
that differs from the model's own answer.  The other models available depend on
when each model was run: Llama-3-8B, Mistral-7B-v0.3, and Qwen2.5-7B draw on
each other, Qwen3.5-9B on these three, and Ministral-3-8B on the other four.
When several candidates exist, we take the one from the model whose name comes
first alphabetically.  When there is none, we supply an off-type answer: the
first gold answer of a random other question, checked to be wrong for this
one.  Off-type answers therefore go to questions where no other model gave a
different wrong answer, mostly questions that the other models answered
correctly.  On these questions the model's own accuracy is $.765$, against
$.236$ on questions with a plausible wrong answer.  A looser normalization
(ignoring case, accents, and plural endings) flags plausible wrong answers that
are close to the reference.  We judged the flagged strings by hand, accepted 26
as correct, and removed the 69 pairs that use them, out of 12,609.

\begin{table}[t]
\footnotesize
\centering
\setlength{\tabcolsep}{3pt}
\resizebox{\columnwidth}{!}{%
\begin{tabular}{@{}lrc@{}}
\toprule
Comparison & $n$ & Difference \\
\midrule
\multicolumn{3}{@{}l}{\emph{Wording of a correct answer (TriviaQA)}} \\
\quad canonical $-$ first alias & 5,000 & $+.114$ \ci{[+.105,+.123]} \\
\quad \quad strings differ & 3,755 & $+.152$ \ci{[+.141,+.164]} \\
\quad \quad strings equal & 1,245 & $-.002$ \ci{[-.006,+.002]} \\
\midrule
\multicolumn{3}{@{}l}{\emph{Right versus wrong}} \\
\quad canonical $-$ plausible wrong & 12,540 & $+.056$ \ci{[+.050,+.062]} \\
\quad \quad wrong answer listed by the benchmark & 412 & $+.083$ \ci{[+.045,+.121]} \\
\midrule
\multicolumn{3}{@{}l}{\emph{Own versus supplied, questions the model got wrong}} \\
\quad own wrong $-$ other model's wrong & 9,359 & $+.044$ \ci{[+.037,+.050]} \\
\quad canonical $-$ own wrong & 9,359 & $-.008$ \ci{[-.015,-.002]} \\
\quad canonical $-$ other model's wrong & 9,359 & $+.036$ \ci{[+.029,+.042]} \\
\midrule
\multicolumn{3}{@{}l}{\emph{Own versus supplied, questions the model got right}} \\
\quad own $-$ canonical, strings differ & 2,904 & $+.005$ \ci{[-.002,+.012]} \\
\bottomrule
\end{tabular}
}
\caption{Paired differences in stated confidence.  Pairs are pooled across the
20 settings, and the 95\% bootstrap intervals resample the pooled pairs.
Strings are compared after normalization (Appendix~\ref{app:repro}).  ``Other model's wrong'' is the plausible wrong
answer; the three rows with $n=9{,}359$ use the same questions.}
\label{tab:app-supplied}
\end{table}

\paragraph{Levels.}  Averaged over the 20 settings, the model states $.913$ for
its own correct answers and $.813$ for its own wrong answers.  It states $.821$
for the canonical reference, $.761$ for a plausible wrong answer, and $.183$
for an off-type answer.  These levels come from different sets of questions,
so Table~\ref{tab:app-supplied} compares answers to the same questions instead.

\paragraph{Wording.}  On TriviaQA, the canonical reference gets more
confidence than the first listed alias.  The gap comes only from questions
where the two strings differ.  Over all questions, it ranges from $+.030$
(Llama-3-8B) to $+.225$ (Qwen3.5-9B) across models.  The canonical form matches the model's
own correct answer in $77\%$ of cases and the first alias in $18\%$.

\paragraph{Right versus wrong.}  The canonical reference beats a plausible
wrong answer by $+.074$ on TriviaQA, $+.074$ on SciQ, $+.076$ on TruthfulQA,
and $+.022$ on MMLU.  The TriviaQA value depends on the wording of the
reference.  If we supply the first listed alias instead, the difference on
TriviaQA becomes $-.019$ ($[-.036,-.002]$): the wrong answer gets more
confidence than the correct one.  On the 412 pairs where the plausible wrong
answer is also a wrong option listed by SciQ, TruthfulQA, or MMLU, and so wrong
by the benchmark's own labels, the margin is $+.083$.

\paragraph{Own versus supplied.}  On the questions the model got wrong, its own
wrong answer gets about as much confidence as the canonical reference and more
than another model's wrong answer.  On the questions it got right, its own
answer and a differently written canonical reference get about the same
confidence.  Re-supplying the model's own answer changes the mean confidence by
at most $.022$ in any setting and leaves $.969$ of the individual values
unchanged on average, so starting the assistant turn with a supplied answer
changes the stated confidence little.

\section{Review of Published Comparisons}
\label{app:audit}

This review asks how published comparisons of verbalized and token confidence
choose and report the protocol.  It is targeted, not a survey: it shows that
the choices differ and are sometimes unstated, and it does not estimate how
often this happens in the field.

\paragraph{Collection and inclusion.}  We collected candidates from web
searches of arXiv and the ACL Anthology on verbalized confidence and its
comparison with token probabilities, from papers suggested to us during
earlier reviewing, and from related work we read during the project.
This gave 22 candidates.  A study is included if it evaluates a verbalized or
self-reported signal and a probabilistic signal tied to one candidate answer,
and states a conclusion about their relative performance or their agreement.
Twelve studies meet this criterion; Table~\ref{tab:audit-exclusions} lists
the ten exclusions.  Every judgment rests on a quote that we located in the
source text, not on a summary of it.

\begin{table*}[t]
\footnotesize
\centering
\setlength{\tabcolsep}{3pt}
\begin{tabular}{@{}p{0.25\linewidth}p{0.69\linewidth}@{}}
\toprule
Excluded candidate & Reason \\
\midrule
\citet{Yang2024}, \citet{Dai2026} &
Study verbalized-confidence elicitation without an evaluated token-score arm. \\
\citet{Wu2026} &
Evaluates self-reported confidence under text-only access; no token-score arm. \\
\citet{YoonKim2025}, \citet{Zeng2025} &
Analyze verbalized confidence across reasoning and non-reasoning models
without an evaluated token-score arm. \\
\citet{Khanmohammadi2025} &
Compares P(True) with a temperature-scaled softmax over the same \texttt{a}/\texttt{b}
label logits of one self-evaluation query, and with trained probes.  The
evaluated signals all derive from one self-evaluation token probability, and
the paper has no verbalized arm in words, so there are not two distinct signal
types. \\
\citet{Ji2025} &
Relates verbal uncertainty to sampling-based semantic entropy, which is tied
to a cluster of sampled meanings rather than to one candidate answer, so the
context and slot axes are undefined. \\
\citet{Geng2024} &
Survey rather than a primary comparison. \\
\citet{Luo2025}, \citet{Xiao2025} &
Propose calibration methods evaluated against baselines, without a stated
conclusion relating a verbalized to a probabilistic-side signal. \\
\bottomrule
\end{tabular}
\caption{Candidates excluded from the targeted protocol audit.  Inclusion
requires an evaluated verbalized or self-reported signal, an evaluated
probabilistic signal tied to one candidate answer---an answer likelihood, an
answer-, label-, or truth-token probability, or a sample-frequency
estimate---and a stated conclusion about their relative performance or their
agreement.  \citet{Xiong2024}, excluded in an earlier reading, is included
(Table~\ref{tab:audit}).}
\label{tab:audit-exclusions}
\end{table*}

Table~\ref{tab:audit} includes studies that evaluate both signal types on the
same items and state a conclusion about their relative performance or their
agreement.  Each row is one method of one study; a study has several rows when
its methods differ in context or aggregation, so the twelve studies give fifteen
rows.  \textbf{S} marks an explicit statement,
\textbf{P} a choice the paper leaves unstated but whose coarse form follows
from the scoring operation the method describes, \textbf{U} no recoverable
choice, and \textbf{--} an inapplicable axis.  The
final column records where the probabilistic-side signal is obtained:
\textbf{G} for a prompt that does not ask for confidence and \textbf{E} for
one that does, whether jointly with the answer or after an answer is
supplied.  The nine studies of the
first round were rated twice by the same rule, and the second pass changed four
judgments, listed after the evidence ledger; the three studies added afterwards
were rated once by that rubric.

\begin{table*}[t]
\scriptsize
\centering
\setlength{\tabcolsep}{3.2pt}
\renewcommand{\arraystretch}{0.95}
\begin{tabular}{@{}llcccccc@{}}
\toprule
Comparison & Probabilistic-side signal & Rated & Scored &
Aggregate & Context/slot & Where & Code \\
\midrule
\citet{LinHiltonEvans2022}, answer logit   & log-probability of the model's answer
  & S & S & U  & P & G & no  \\
\citet{LinHiltonEvans2022}, indirect logit & log-probability of ``True'' after the answer
  & S & S & -- & S & E & no  \\
\citet{Kadavath2022}         & P(True) token probability
  & S & S & -- & S & E & no  \\
\citet{Tian2023}             & label probability, $n{=}10$ sample frequency
  & S & P & -- & P & G & no  \\
\citet{Ni2024}               & mean negative log-probability of the answer
  & S & S & S  & S & E & no  \\
\citet{Kumar2024}            & softmax over answer-option tokens
  & S & S & S  & S & G & yes \\
\citet{Lyu2024}              & $\exp$(mean log $p$) $=1/\mathrm{PPL}$
  & S & S & S  & P & G & no  \\
\citet{Huang2024}            & sequence likelihood, length-normalized
  & S & S & S  & S & G & yes \\
\citet{Xiong2024}            & agreement of sampled answers
  & S & P & -- & S & E & yes \\
\citet{Tao2025}              & inverse perplexity of the generated response
  & S & P & S  & S & E & no  \\
\citet{Xia2026}              & length-normalized token probability
  & S & S & P  & S & G & no  \\
\citet{Testoni2026}          & mean token probability; G-NLL
  & S & S & S  & S & G & yes \\
\citet{Muller2026}, P(True)  & truth-label probability (Exp.\ 2)
  & S & S & -- & P & E & yes \\
\citet{Muller2026}, CCP      & product of token confidences (Exp.\ 2)
  & S & S & S  & S & G & yes \\
\citet{Muller2026}, consistency & agreement of 10 samples (Exp.\ 2)
  & S & S & -- & S & G & yes \\
\midrule
\multicolumn{2}{@{}l}{\emph{explicitly specified}}
  & 15/15 & 12/15 & 7/15 & \textbf{11/15} & 6 G / 5 E & 7/15 \\
\multicolumn{2}{@{}l}{\emph{incl.\ reconstruction}}
  & -- & -- & -- & -- & 9 G / 6 E & -- \\
\bottomrule
\end{tabular}
\caption{Specification audit of published verbalized-vs-probabilistic
confidence comparisons.  \textbf{S}: the paper states the choice; \textbf{P}: it does not,
but the scoring operation the method describes fixes the coarse context;
\textbf{U}: not recoverable; \textbf{--}: not applicable.  \textbf{Where}
records the context the probabilistic-side signal is obtained under: \textbf{G} an
answer-generation prompt that does not ask for confidence, and \textbf{E} a
prompt that asks for confidence, either jointly with the answer or after an answer is
supplied.  The first summary row counts \textbf{S} only, so its \textbf{Where}
entry covers the eleven rows that state the choice; the second adds the four
that do not, reconstructing each from its own scoring operation, the
answer-generating prompt for the answer-likelihood methods and the confidence
query in which the truth token is scored for the P(True) methods.  Counts are over the fifteen rows; seven rows, from five of the twelve studies,
link code.  A
\textbf{P} or \textbf{U} records under-specification, not an invalid protocol
or result.}
\label{tab:audit}
\end{table*}

Table~\ref{tab:audit-evidence} records the source locations used to assign
the non-\textbf{U} cells.  For context/slot, we searched the methods and
prompt appendices for both the conditioning prompt and the answer position.
A paper receives \textbf{S} only when both are determined.  Code availability
means an author-linked implementation of the evaluated method, not merely a
model or dataset repository.

\begin{table*}[t]
\scriptsize
\centering
\setlength{\tabcolsep}{3.2pt}
\begin{tabular}{@{}lp{0.67\linewidth}@{}}
\toprule
Comparison & Evidence used for the audit judgments \\
\midrule
\citet{LinHiltonEvans2022}, answer logit &
Abstract states that the model generates an answer and confidence; \S2.1
defines answer and indirect logits as ``based on the log-probabilities that a
language model assigns to tokens''.  Figure~2 shows the answer logit as the
normalized log-probability of the model's own zero-shot answer.  CalibratedMath
answers ``are always integers'', so a multi-token answer is possible and no
aggregation rule is given.  The logit is read on the model's own answer, which
fixes the context only by that convention. \\
\citet{LinHiltonEvans2022}, indirect logit &
Figure~2 shows the model's greedy answer appended to the question, followed by
a true/false query, with the score read as the ``Logprob for `True' token''.
The scored unit is one token, so aggregation does not apply, and the appended
query is the conditioning text, which places this method in \textbf{E}. \\
\citet{Kadavath2022} &
Abstract and \S4 describe first proposing an answer and then evaluating
P(True).  \S4 conditions the single True-token probability on the question
and proposed answer, fixing the rated answer, scored answer, and context;
span aggregation is inapplicable. \\
\citet{Tian2023} &
Table~1 distinguishes one- and two-stage elicitation.  \S2 defines Label
probability as $p(y\mid x)$ estimated from ten samples because per-token
probabilities are unavailable.  The label and bare-question context are
inferable but the alignment with the verbalized answer slot is not explicit. \\
\citet{Ni2024} &
Abstract and \S1 compare probabilistic and verbalized perception and conclude
that ``for less common questions, probabilistic perception outperforms
verbalized perception''.  \S3.1 defines $c_{\mathrm{prob}}$ as
$-\frac{1}{n}\sum_i \log P(a_i\mid a_{<i})$ over the $n$ answer tokens; the
formula suppresses the prompt, but \S3.2 gives the single prompt verbatim, and
it ends in \texttt{Answer:} and asks for a \texttt{certain}/\texttt{uncertain}
marker after the answer, so the conditioning text and the answer position are
both determined.  That prompt asks for confidence, so the score is read in
an elicitation context.  No repository is linked. \\
\citet{Kumar2024} &
The paper evaluates token-probability confidence and verbalized certainty on
the same items and concludes about their alignment, reporting the strongest
alignment for GPT-4, so both arms are evaluated and compared.
\S2.1 gives the serialized prompt, ``$Q$ + newline + options + newline +
\texttt{Answer: }'', and \S2.2 derives internal confidence from a softmax over
the generated answer-option tokens with an explicit adjustment procedure, so
the context, the position, and the aggregation are all determined.  The paper
links reproduction code. \\
\citet{Lyu2024} &
Methods define the raw-logit score as the exponential of the average log
probability of all tokens in the sampled reasoning chain, equivalently
inverse perplexity.  No token-score conditioning prompt or position is stated. \\
\citet{Huang2024} &
\S2 defines the negative log-likelihood of a generated response and its
length-normalized variant.  \S3 states, ``We use the same prompt template as
that in Lin et al.\ (2023),'' and Appendix~E lists the generation prompts, so
the score is read on the response under its generation prompt.  Verbalized
confidence is elicited separately, ``by feeding a query-response pair to an
LM''.  The abstract links code to replicate the experiments. \\
\citet{Xiong2024} &
Figure~1 gives the prompt ``Provide the answer and your confidence in the
answer.''  \S3.4 measures confidence as the agreement between sampled
candidate answers and the original answer, so the scored string is inferable
but not stated, and token aggregation does not apply.  The Table~3 caption
states, ``The prompt and aggregation strategies are fixed as CoT and
Consistency when M > 1,'' and the CoT prompt asks for verbalized confidence.
\S5.3 states that ``the sampling strategy with 5 sampled responses paired with
consistency aggregation consistently outperform verbalized confidence''.  The abstract links code. \\
\citet{Tao2025} &
Abstract compares token-probability, numerical-verbal, and linguistic-verbal
uncertainty.  The introduction (PDF p.~2) describes the token-probability method
as one that ``estimates uncertainty from token-level probabilities using the
inverse of perplexity over the generated response'', which fixes the scored
string and the aggregation.  \S2 states, ``To enable consistent and fair
uncertainty extraction across methods, we adopt a unified prompt template that
combines numerical verbal uncertainty and linguistic verbal uncertainty formats
\ldots\ Detailed prompt formats are provided in Appendix~F.''  The token arm is
therefore read under a prompt that also elicits both verbal formats. \\
\citet{Xia2026} &
\S2 defines linguistic confidence and self-evaluation; the evaluated
probability method ``aggregates the token probabilities of the generated
answer under greedy decoding and normalizes them by length'', which fixes the
scored string and the context but not the exact aggregation function.
Appendix~A.3 states, ``To evaluate robustness, we estimate confidence scores
for the same answer under different prompt formulations,'' and notes that
missing GPT-4o logits prevent ``computing the probability of a fixed answer
across prompts''.  This is a same-answer cross-prompt measurement, although
the compared prompts are task-prompt variants rather than a generation and a
confidence-elicitation context. \\
\citet{Testoni2026} &
\S5.1 compares logit, elicitation, and consistency methods; \S4.1 defines
average token-level probability under greedy decoding and G-NLL, and states
that models with a recommended system prompt use it as provided by the model
developers.  Appendix~B (``Model Responses'') gives the generation prompt,
which asks to ``Provide a concise answer and start your response with the
letter of the selected option in square brackets'' and ends in ``Your Answer:''.
The score is read on the response under that prompt, so the context and the
answer position are determined.  The introduction links the code. \\
\citet{Muller2026} &
We audit the arXiv v2 text (``Semantic Calibration Prevails Where Token
Confidence Fails''); the implementation details are in v2 Appendix~A.11.3 and in
v1 Appendix~A.10.2 (v2 A.10.3 is ``Selection Bias'').  Experiment~1 scores a
single label-token probability and does not involve verbalized confidence, so
we do not use it, nor its ``Prompt Design 1'', for this audit.  Experiment~2
(\S3.5) ``evaluates sequence-level UQ calibration'' with four methods
``applied to identical cached long-form generations'': verbalized
uncertainty, P(True), claim-conditioned probability (CCP), and answer
consistency.  The verbalized prompt is adopted from \citet{Tian2023}.
\emph{P(True)}: the prompt ``follows the formulation of Kadavath et al.\
(2022)'' and the exact prompts are given in the code repository; the signal is
the probability of the truth label, so aggregation does not apply, and the
context follows from the formulation, hence \textbf{P}, in \textbf{E}.
\emph{CCP}: ``aggregating token-level confidence scores through multiplicative
composition'', read on the cached response, whose question-answering prompts
(APriCoT for multiple choice, chain of thought otherwise) are listed in
Appendix~A.11.1.  \emph{Answer consistency}: the share of semantically
equivalent answers among 10 samples generated under the same prompts;
aggregation does not apply.  The paper links its benchmark implementation. \\
\bottomrule
\end{tabular}
\caption{Evidence ledger for Table~\ref{tab:audit}.  Locations were checked
against the paper text.  An absent context/slot statement is not evidence that
an author used an invalid protocol; it means only that a reimplementation must
choose the context and output position.}
\label{tab:audit-evidence}
\end{table*}

\paragraph{Second rating pass.}
Because the audit rests on one rater, we derived it a second time.  For the
nine studies of the first round we re-downloaded every source, re-located every
quoted string, and, for the context axis, extracted every sentence mentioning
the probabilistic-side signal and read those that also mention a prompt,
context, or position.
All quoted strings were found again.  Four judgments changed: \citet{Tao2025} and \citet{Muller2026}
from unspecified to stated, \citet{Xia2026} from unspecified to stated for
context and partly determined for aggregation, \citet{LinHiltonEvans2022} from
inapplicable to unspecified for aggregation, and the aggregation evidence for
one further study, which had been taken from that paper's description of a
different method.  A later refinement of the \textbf{P} category moved the
context axis of \citet{LinHiltonEvans2022}, \citet{Lyu2024},
and \citet{Testoni2026} from unspecified to partly determined, and that of \citet{Huang2024} to stated once we located
its generation prompt.  The first pass had applied two different questions:
whether a reimplementation can determine the probabilistic-side context, and whether
the paper says the two signals share one context.  The table uses the first
question throughout.  \citet{Ni2024} and \citet{Kumar2024} were added after
that round, and \citet{Xiong2024} after a later re-reading of the inclusion
criterion; each was rated by the same rubric in a single pass over its source;
for \citet{Ni2024} the context axis was re-derived once more after an initial
reading treated the suppressed prompt in its scoring formula as an omission
rather than as shorthand.

\paragraph{Source corrections.}
A last comparison against the source texts, for the arXiv versions named in the
ledger, changed the following.  The aggregation of \citet{Tao2025} moved from
unspecified to stated, because the introduction describes the inverse perplexity
of the generated response.  The code of \citet{Huang2024} and
\citet{Testoni2026} is linked in the abstract and the introduction.  The context
of \citet{Testoni2026} moved from partly determined to stated, because
Appendix~B gives the generation prompt and how the response begins.
\citet{LinHiltonEvans2022} is split into the answer logit, read on the model's
answer, and the indirect logit, read after an appended true/false query.
\citet{Muller2026} is rated on its second experiment, the one that compares
verbalized confidence with other signals, with its methods in separate rows.
\citet{Khanmohammadi2025} is excluded, because its compared signals all derive
from one self-evaluation token probability and it has no verbalized arm in
words.  These changes were made in a single pass over each source, by the same
rubric.

The distinction matters in practice.  \citet{Tian2023} use a ten-sample
frequency estimate rather than per-token likelihood, so an open-weight
reimplementation has no original token score or answer slot to copy.
\citet{Yang2024} likewise identify logit aggregation as a design choice, and
\citet{Xia2026} note that prompt robustness remains understudied.  These points
motivate explicit specification; they do not invalidate the cited results.

\end{document}